\documentclass{article}

\usepackage{arxiv}
\usepackage{amsmath}
\usepackage{amssymb}
\usepackage{amsfonts}

\usepackage[utf8]{inputenc} 
\usepackage[T1]{fontenc}    
\usepackage{hyperref}       
\usepackage{url}            
\usepackage{booktabs}       
\usepackage{amsfonts}       
\usepackage{nicefrac}       
\usepackage{microtype}      
\usepackage{cleveref}       
\usepackage{lipsum}         
\usepackage{graphicx}
\usepackage{natbib}
\usepackage{doi}

\DeclareMathOperator*{\argmin}{arg\,min}

\usepackage{graphicx}
\usepackage{url}
\usepackage{multirow}
\usepackage{subfigure}
\usepackage{booktabs}

\title{Integrating Nearest Neighbors with Neural Network Models for Treatment Effect Estimation}

\author{ \href{https://orcid.org/0000-0003-1729-4124}{\includegraphics[scale=0.06]{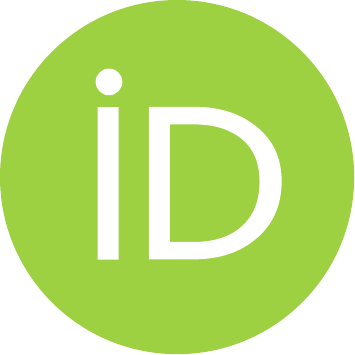}\hspace{1mm}Niki Kiriakidou}\thanks{Corresponding author.} \\
	Department of Informatics and Telematics,\\
	Harokopio University of Athens,\\
	Athens, GR177 78.\\
	\texttt{kiriakidou@hua.gr} \\
	\And
	\href{https://orcid.org/0000-0002-2461-1928}{\includegraphics[scale=0.06]{orcid.pdf}\hspace{1mm}Christos Diou} \\
	Department of Informatics and Telematics,\\
	Harokopio University of Athens,\\
	Athens, GR177 78.\\
	\texttt{cdiou@hua.gr} \\
}

\renewcommand{\shorttitle}{Integrating Nearest Neighbors with Neural Network Models for Treatment Effect Estimation}

\hypersetup{
pdftitle={Integrating nearest neighbors on neural network models for treatment effect estimation},
pdfsubject={q-bio.NC, q-bio.QM},
pdfauthor={Niki Kiriakidou, Christos Diou},
pdfkeywords={Causal inference, neural networks, average treatment effect, individual treatment effect, counterfactual},
}

\begin{document}
\maketitle

\begin{abstract}
Treatment effect estimation is of high-importance for both researchers and practitioners across many scientific and industrial domains. The abundance of observational data  makes them increasingly used by researchers for the estimation of causal effects. However, these data suffer 
from several weaknesses, leading to inaccurate causal effect estimations, if not handled properly. 
Therefore, several machine learning techniques have been proposed, most of them focusing on leveraging the predictive power of neural network models to attain more
precise estimation of causal effects. 
In this work, we propose a new methodology, named Nearest Neighboring Information for Causal Inference (NNCI), for integrating valuable nearest neighboring
information on neural network-based models for estimating treatment effects. The proposed NNCI methodology is applied to some of the most well established neural network-based
models for treatment effect estimation with the use of observational data.
Numerical experiments and analysis provide empirical and statistical evidence that the
integration of NNCI with state-of-the-art neural
network models leads to considerably improved treatment effect estimations on a
variety of well-known challenging benchmarks.
\end{abstract}


\section{Introduction}\label{sec1}

Causal effect estimation is often the central research data analysis objective
across many scientific disciplines.
{ In particular, it is the process of inferring a causal relationship
	between an intervention, or else treatment, and its effect on an outcome variable of interest.
	It involves quantifying the} extent to which a change in { a 
	given intervention or treatment} would influence the variable of interest.
Among others, treatment effect is used in healthcare \cite{schneeweiss2009high},\cite{zheng2022kuramoto},
education \cite{oreopoulos2006estimating} and advertising
\cite{zantedeschi2017measuring}.
{Some} examples of causal questions in each {one} of these fields 
include 
``\textit{What is the effect of a specific drug on a patient's blood pressure levels?}'', 
``\textit{What is the effect of studying two more hours on a student's final test performance?}'', 
``\textit{What is the effect of an advertisement on social media to the product's sales?}'', respectively.

The golden standard for answering these questions is through the conduction of
Randomized Controlled Trials (RCTs), in which subjects are randomly assigned to
two groups: the \textit{treatment group} (also known as intervention or experimental)
and the \textit{control group}. The intervention of interest is applied on the members
of the treatment group, while no intervention is applied on the members of the
control group. Due to randomization, { RCTs enable the calculation of the real treatment} effect.
Nevertheless, in most of the cases it is impossible to conduct an RCT due to
financial and/or ethical issues, or due to the large number of combinations of
variables that need to be evaluated. 

{Therefore, researchers from various high-impact scientific areas could be benefited from the plethora of readily available observational data for the estimation of treatment effects. Nevertheless, handling observational data enclose several complications \cite{kuang2020causal,hammerton2021causal}, since actions and outcomes are observed retrospectively and the mechanism caused the action is unknown. Additionally, observational data may include possible (hidden or observed) counfounding variables, which lead to incorrect estimation of treatment effects.
	Hence, there has been a growing interest in using machine learning models for causal effect estimation, as these models can leverage observational data as well as they can
	capture complex relationships between variables and provide accurate predictions.}

In this work, we propose a new methodology for integrating information from
neighboring samples in neural network architectures for treatment effect
estimation. 
The proposed methodology, {named Nearest Neighboring Information for Causal Inference (NNCI), is implemented on the most effective
	neural network-based models, which provide predictions of the outcomes from control and treatment groups as well as for the propensity score, i.e., the probability for  a subject to be assigned into the treatment group. The motivation of our approach consists of enriching the models' inputs with valuable information from control and treatment groups along with the covariates in order to improve the estimations of treatment effects. The proposed methodology identifies the nearest neighbor instances in the control group and the treatment group for each instance 
	and then calculates the average outcomes of identified instances contained in both groups. This information is integrated with the covariates to the model's inputs in order to increase the prediction accuracy and reduce bias.
	By using the outcomes of the nearest neighbors as input features, the methodology aims to capture the causal effects of the treatment more efficiently. The neural network models can learn to weight these new input features appropriately, and better capture complex relationships between the features and the treatment effect.
	Summarizing, the main contributions of this work are:}
\begin{itemize}
	
	\item {NNCI methodology for integrating information from neighboring instances in neural network-based causal inference models for treatment effect estimation.}	
	\item the modification of the architecture of three state-of-the-art neural network models i.e., Dragonnet \cite{shi2019adapting}, TARnet \cite{shalit2017estimating} and NEDnet \cite{shi2019adapting} {for incorporating the information provided by NNCI}. {The developed models} use a vector of features extracted from the outcomes of the nearest neighbors of each sample, separately for the treatment and control group as inputs, to improve treatment effect estimation.
	\item a comprehensive experimental analysis based on
\cite{dolan2002benchmarking} performance profiles as well as on post-hoc and non-parametric statistical tests \cite{finner1993monotonicity,hodges1962rank}. The conducted analysis demonstrate that in most cases the proposed methodology leads to considerable improvement in the estimation of treatment effects, while in a minority of the cases, it exhibits similar performance with the corresponding baseline model.
\end{itemize}
%

The remainder of this paper is organized as follows:
Section \ref{sec:related_work} presents a brief review of the
state-of-the-art baseline models for treatment effect estimation.
Section \ref{sec:methodology} presents a detailed description of the proposed methodology and the {proposed} neural network-based models in which NNCI methodology {is adopted}.
Section \ref{sec:evaluation} presents the used datasets and the evaluation methodology.
Section \ref{sec:experiments} presents the numerical
experiments, focusing on the evaluation of the proposed {neural network-based causal inference models} with the corresponding baseline models.
{Section \ref{sec:discussion} discusses the proposed framework
	as well as the experimental results and outlines the findings of this work.}
Finally, Section \ref{sec:conclusions} summarizes the main findings and conclusions of this work as well as some interesting directions for future work. 

\section{Related Work}\label{sec:related_work}

In recent years, researchers aimed at leveraging machine learning models for treatment
effect estimation from observational data.
Neural networks have demonstrated that they offer high capacity, while at the
same time avoiding overfitting for a range of applications, including
representation learning. Therefore, there has been considerable interest in
using them for the estimation of causal effects at both the individual and
population level. Next, we briefly describe the most well-mentioned.

Yoon et al. \cite{yoon2018ganite} proposed 
	a new causal inference model, named Generative Adversarial Nets for the inference of Individualized Treatment Effects (GANITE) for estimating individual treatment effects. 
	The rationale of their approach was the simulation of uncertainty regarding the counterfactual distributions, which was achieved by learning these distributions using a GAN model. 
	The presented numerical experiments revealed that GANITE outperformed
	S-learner machine learning models \cite{kunzel2019metalearners}
	as well as some tree-based models (BART \cite{chipman2010bart}, R-Forest \cite{breiman2001random} and C-Forest \cite{wager2018estimation}) on three benchmarks.

 \cite{louizos2017causal} proposed Causal Effect Variational Auto-Encoder (CEVAE), for leveraging the proxy variables for the accurate
	estimation of treatment effects. The main advantage of the CEVAE is that it requires
	substantially weaker assumptions about the structure of the hidden confounders 
	as well as about the data generating process.
	The authors provided a comprehensive experimental analysis, which
	presented that CEVAE exhibited better performance compared to traditional
	causal inference models and presented more robust behavior against
	hidden confounders in the case of noisy proxies.

\cite{shalit2017estimating} proposed a new framework, named Counterfactual Regression (CFR), which focuses on learning a balanced representation of treatment and control groups using a prediction model. The authors utilized two distances, namely Maximum Mean Discrepancy (MMD) \cite{gretton2012kernel} and Wasserstein distance (Wass) \cite{villani2009optimal} to measure the distances between treatment and control groups' distributions. In addition, they proposed Treatment Agnostic Representation Network (TARnet) neural network model, which constitutes a variant of CFR without balance regularization.
The experimental analysis revealed the superiority of TARnet and CFR (Wass) over 
S-learner and tree-based causal inference models.

Recently, \cite{shi2019adapting} proposed a three-head neural network model, named
Dragonnet for the estimation of conditional outcomes as well as the propensity score and a new loss function, named \emph{targeted regularization} for further improving the estimation of causal effects and reduce the bias of the estimator. Furthermore, the authors proposed a modification of Dragonnet, named NEDnet, which was trained with a multi-stage procedure instead of an end-to-end. The main difference is that NEDnet is first trained using a pure treatment prediction objective, which is then replaced  with an outcome-prediction matching (three-head), similar to the one used by Dragonnet. The representation layers are then frozen and the outcome-prediction neural network is trained on the pure outcome prediction task.
The reported experimental results showed that both Dragonnet and NEDnet achieved to get better estimations than the state-of-the-art models and concluded that both models along with targeted regularization substantially improve estimation quality.

Kiriakidou and Diou \cite{KiriakidouAIAI22} proposed a neural network
causal inference model, named modified Dragonnet. 
The proposed model captures information not only from covariates, but also from the average outcomes of neighboring instances from both treatment and control groups.For evaluating the efficiency of their approach, they used the semi-synthetic collection datasets IHDP \cite{hill2011bayesian}. In their experiments, the proposed model was implemented with three different Minkowski distance metrics for the calculation of	neighboring instances. The presented  experimental analysis revealed that modified Dragonnet constitutes a better estimator than Dragonnet for all utilized metrics, while simultaneously is able to predict treatment effects with high accuracy. Nevertheless, the limitation of this approach was that the authors adopted the proposed approach on one neural network-based causal inference model and that the evaluation was based only on one benchmark.

In this research, we present an extension of our previous work \cite{KiriakidouAIAI22} by proposing a new methodology, {named Nearest Neighboring Information for Causal Inference (NNCI)}, based on the exploitation of valuable information provided by nearest neighboring {instances-based philosophy for each instance}. {NNCI is applied on the most well-established neural network-based causal inference models proposed in the literature, namely
	Dragonnet, TARnet and NEDnet for capturing the causal effect estimations more accurately.
	Notice that these models were selected due to their special architecture design and the fact that they constitute the only 
	neural network-based model to provide estimations for the propensity score as well as the conditional outcomes for control and treatment groups.
	The major difference between the presented works and the proposed approach is that the former ignore valuable information contained in the outcomes of the training instances; while the latter enriches the models' inputs with information from the covariates as well as from the outcomes from the control group and the treatment group.
	The main idea of the proposed approach is that the neural network causal inference models, through the adoption of NNCI are able to better capture complex causal relationships between the features and the treatment effect
	by appropriately weighting the advanced input features.}
Furthermore, in contrast to the usual approach for the
performance evaluation of models for treatment effect estimation 
\cite{yao2018representation,shi2019adapting,johansson2016learning,shalit2017estimating,louizos2017causal}, we provide concrete and empirical evidence about the superiority of our approach, by using Dolan and Mor{\'e}'s \cite{dolan2002benchmarking} performance profiles and a detailed
statistical analysis based on non-parametric and post-hoc statistical tests \cite{hodges1962rank,finner1993monotonicity}.

\section{Methodology}\label{sec:methodology}

In this section, we provide a detailed presentation of the proposed methodology for treatment effect estimation. {We recall that the rationale behind our approach is to exploit the wealth of information from nearest neighboring instances from the training data in order to get more accurate estimations of average and individual treatment effects. 
	This is achieved through the enrichment of model's inputs with the average outcomes of the nearest neighbors from the control and treatment groups of each instance contained in the training data.}

\subsection{Problem setup}

In this work, we are relying on the potential outcome framework of Neyman-Rubin
\cite{rubin2005causal}. We consider a $d$-dimensional space of covariates $X$
and a joint distribution $\Pi$ on $\mathcal{X}\times \mathcal{T} \times
\mathcal{Y}$, where $\mathcal{X}$, $\mathcal{T}$ and $\mathcal{Y}$ are the
domains of random variables corresponding to the covariates, the treatment and
the outcome of a sample, respectively.

According to the potential outcome framework we can learn causal effects given on
a set of treatment-outcome pairs ($T, Y$). Throughout the paper, we are
considering the case of binary treatment, which means that $T\in \{0,1 \}$, where
$T=0$ and $T=1$ corresponds to the samples belong to the control and
treatment group, respectively. Notice that for every instance $i$, there is a
potential outcome $y_{i}^{(T)}$: $y_{i}^{(0)}$ for the samples belong to the control group and $y_{i}^{(1)}$ for the samples belong to the
treatment group. It is worth mentioning that the fundamental problem of causal inference is that only one
of the potential outcomes can be observed for each instance. In case the
sample belongs to the treatment group, then $y_{i}^{(1)}$ is the factual
outcome and $y_{i}^{(0)}$ is the counterfactual outcome, and vice-versa for
samples belong to the control group.

Treatment effects can be defined using the potential outcomes $y_{i}^{(0)}$ and
$y_{i}^{(1)}$. One of the causal effects we are interested to estimate is the
\emph{Individual Treatment Effect} (ITE), which measures the effect of the
treatment on a single sample:
\begin{equation}
	\text{ITE}_{i} = y_{i}^{(1)} - y_{i}^{(0)}
\end{equation}
To measure the causal effect of a treatment over the whole
population, we use the \emph{Average Treatment Effect} (ATE):
\begin{equation}
	\text{ATE} = E[Y\,\vert \, T=1]- E[Y\,\vert \, T=0]
\end{equation} 
{Finally, for measuring} the treatment effect for a specific
subgroup of the population, i.e., samples with a specific value to the
covariates $X$, we estimate the \emph{Conditional Average Treatment Effect}
(CATE):
\begin{equation}
	\hspace{-.2cm}  \text{CATE}(\textbf{x}) = E[Y\,\vert \, X=\mathbf{x} , T=1] - E[Y\,\vert \, X=\mathbf{x}, T=0]
\end{equation}

A standard dataset for inferring treatment effect consists of the covariate
matrix $\mathbf{X}$, the treatment vector $\mathbf{t}$ and the outcome vector
$\mathbf{y}$.

\subsection{Nearest neighboring information for causal inference}


The proposed Nearest Neighboring Information for Causal Inference (NNCI)
methodology aims to enrich the inputs of neural network models with information from nearest neighboring instances, to estimate treatment effects from observational data. {More specifically, for each instance $\mathbf{x}_i$, NNCI identifies its $k$ nearest neighboring instances in the control and treatment groups. Then, we calculate the average outcomes of the identified instances from both groups. These quantities, denoted as $\overline{y}_{i}^{(0)}$ and $\overline{y}_{i}^{(1)}$, are incorporated as additional input features to the neural network models. By using the outcomes of the nearest neighbors $\overline{y}_{i}^{(0)}$ and $\overline{y}_{i}^{(1)}$ as input features, the NNCI methodology aims to capture the effect of the intervention more efficiently. The neural network models can learn to weight these new input features appropriately; hence, to better capture complex relationships between the features and the treatment effect.}

\begin{figure}[ht]
	\centering
	\includegraphics[width=10cm]{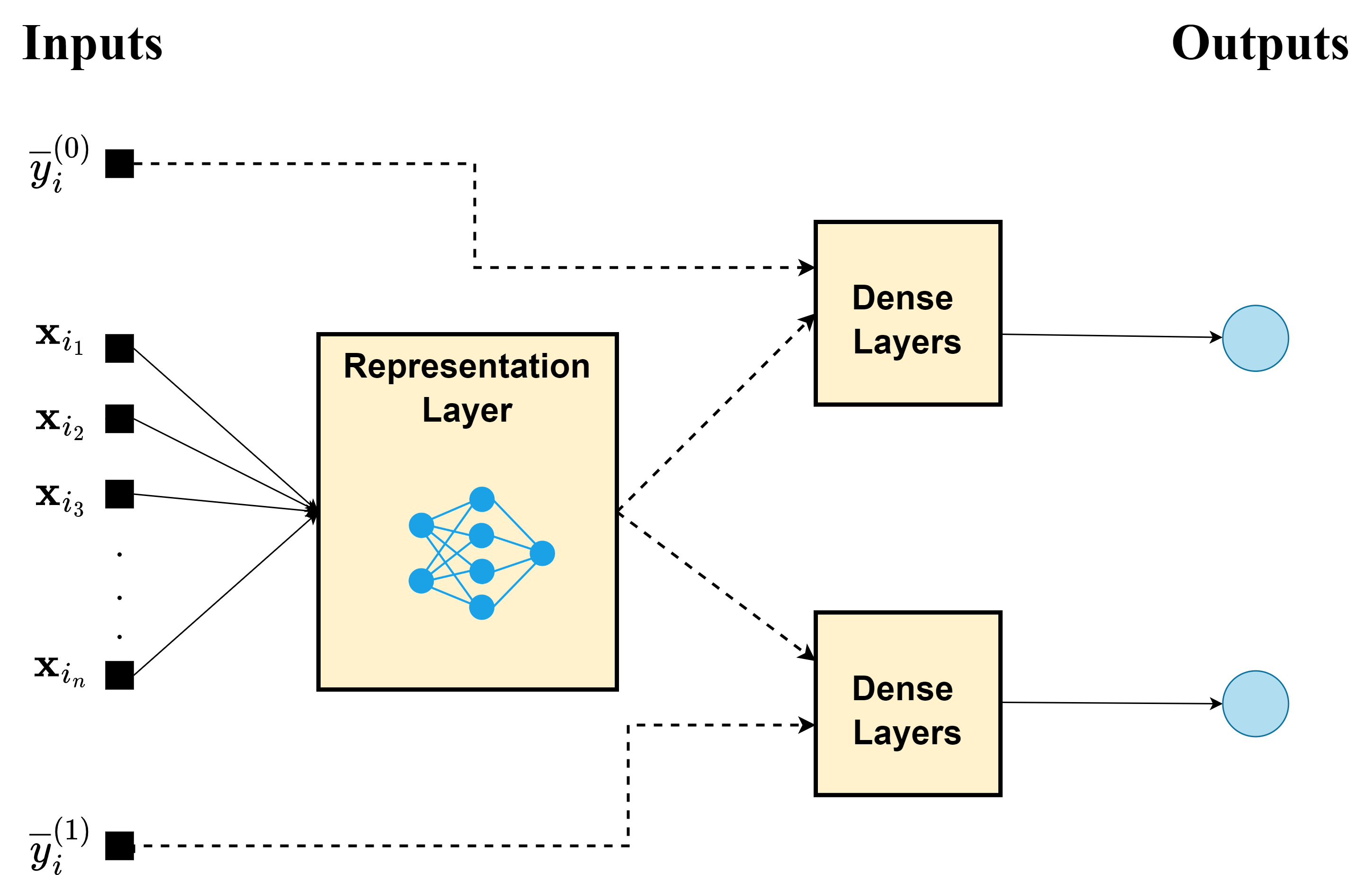}
	\caption{Example of the application of NNCI methodology to a neural network model}\label{Fig:NNCI}
\end{figure}

{NNCI is integrated to the most effective neural network-based causal
	inference models i.e. Dragonnet, TARnet and NEDnet due to their special architecture design. 
	More analytically, these models use a deep net to produce a representation
	layer, which is then processed independently for predicting both the outcomes for treatment and control groups.}
The adoption of NNCI to these models has the result of using as inputs the instance 
$\mathbf{x}_i {= (\mathbf{x}_{i_1},\mathbf{x}_{i_2}, \cdot \cdot \cdot, \mathbf{x}_{i_n})}$ contained in the covariate matrix $\mathbf{X}$ along with the average outcomes of the neighboring instances from treatment and control groups,
$\overline{y}_{i}^{(1)}$ and $\overline{y}_{i}^{(0)}$, respectively, as it is
presented in Figure~\ref{Fig:NNCI}.

Algorithm 1 presents the pseudocode of NNCI methodology for calculating
information from the outcomes of the instances contained in the training
data. The inputs are the selected number of nearest neighbors $k$, the
covariate matrix $\mathbf{X}$, the binary vector of treatment values
$\mathbf{t}$ and the vector with the outcome values $\mathbf{y}$. 
The outputs are the vectors $\overline{\mathbf{y}}^{(0)}$ and $\overline{\mathbf{y}}^{(1)}$, containing the
average values of neighboring outcomes for each sample from control and treatment group,
respectively.

In Step 1, $\mathbf{y}^{(0)}$ and $\mathbf{y}^{(1)}$ are set to $\mathbf{0}$. 
For every instance $\mathbf{x}_{i}$, NNCI calculates the average values of
neighboring outcomes from the control and treated groups (Steps 2-7).
In more detail, in Step 4, NNCI calculates the $k$-nearest neighbors of instance
$\mathbf{x}_{i}$, which belong in the control group (i.e $T=0$) and store their
corresponding indices in the index set $S_0$.
In Step 5, NNCI calculates the mean value of these neighboring outcomes, $\displaystyle
\overline{y}_i^{(0)} = \frac{1}{k} \sum_{j\in S_0} y_j $.
Similarly, in Step 6-7, NNCI calculates the mean value of $k$-nearest neighbors'
outcomes in treatment group (i.e $T=1$), namely $\overline{y}_i^{(1)} =
\frac{1}{k} \sum_{j\in S_1} y_j$.

\vspace{2cm}

\noindent{}\textbf{Algorithm 1: NNCI}

\noindent\rule{8cm}{0.4pt}\vspace{-.3cm}
\begin{description}
	\item[Inputs:]
	\item \quad $k$: number of nearest neighbors
	\item \quad $\mathbf{X}$: covariate matrix      
	\item \quad $\mathbf{t}$: vector of treatment values $t$
	\item \quad $\mathbf{y}$: vector of outcome values $y$ \vspace{.2cm}
	
	\item[Outputs:]
	\item \quad $\overline{\mathbf{y}}^{(0)}$: vector with average of $k$-nearest outcomes from control group for each sample
	\item \quad $\overline{\mathbf{y}}^{(1)}$: vector with average of $k$-nearest outcomes from treatment group for each sample
	
	\item[]
	\setlength\itemsep{.3em}
	\item[Step 1:] Set  $\overline{\mathbf{y}}^{(0)} = \mathbf{0}$ and $\overline{\mathbf{y}}^{(1)} =\mathbf{0}$
	\item[Step 2:] \textbf{for} $i=1$ \textbf{to} $n$ \textbf{do}
	\item[Step 3:] \hspace{.4cm} $\mathbf{x}_{i}=\mathbf{X}[i,:]$
	\item[Step 4:] \hspace{.4cm} Calculate the index set $S_0$ containing the\vspace{-.1cm}
	\item[] \hspace{1.6cm}indices of the $k$-nearest neighbors of $\mathbf{x}_{i}$\vspace{-.1cm} 
	\item[] \hspace{1.6cm}with $T = 0$
	\item[Step 5:] \hspace{.4cm} $\displaystyle \overline{y}_i^{(0)} = \frac{1}{k} \sum_{j\in S_0} y_j $
	%
	
	%
	\item[Step 6:] \hspace{.4cm}Calculate the index set $S_1$ containing the\vspace{-.1cm}
	\item[] \hspace{1.6cm}indices of the $k$-nearest neighbors of $\mathbf{x}_{i}$\vspace{-.1cm} 
	\item[] \hspace{1.6cm}with $T = 1$
	\item[Step 7:] \hspace{.4cm} $\displaystyle\overline{y}_i^{(1)} = \frac{1}{k} \sum_{j\in S_1} y_j $
	\item[Step 8:] \textbf{end}			
	\item[Step 9:] Return $\overline{\mathbf{y}}^{(0)}$, $\overline{\mathbf{y}}^{(1)}$.
\end{description}\vspace{-.5cm}
\noindent\rule{8cm}{0.4pt}\\

Based on this iterative process, NNCI calculates the average outcomes from control
and treated groups and store these quantities into $\overline{\mathbf{y}}^{(0)}$
and $\overline{\mathbf{y}}^{(1)}$, respectively. These vectors are then used in
conjuction with Dragonnet, TARnet and NEDnet models for the prediction of
conditional outcomes $Q(t,\mathbf{x}) = E(Y\, |\, X =\mathbf{x}, T=t)$ and
propensity score $g(x) = P(T = 1\, |\, X =\mathbf{x})$, which measures the
probability of a subject to belong in the treatment group, based on its
characteristics.
{The computational complexity for calculating the vectors $\overline{\mathbf{y}}^{(0)}$ and  $\overline{\mathbf{y}}^{(1)}$ is $O(n^2)$, where $n$ is the number of training instances. This implies that for large datasets, the computational cost may generally be considered high.
	In these cases, one can use a sub-sampling strategy, where only a randomly selected subset of the training set is considered for the estimation of vectors $\overline{\mathbf{y}}^{(0)}$ and  $\overline{\mathbf{y}}^{(1)}$ and the cost would be $O(m \cdot n)$, where $m << n$ is the number of samples used for the estimation.}

\subsubsection{NN-Dragonnet}

Next, we present the NN-Dragonnet model, which is based on the adoption of the
NNCI methodology in the state-of-the-art Dragonnet model
\cite{shi2019adapting}. The inputs of NN-Dragonnet model are the instance
$\mathbf{x}_i$, the average outcomes of its $k$ neighboring instances from the
control group $\overline{y}^{(0)}_i$ and the average outcomes of its $k$
neighboring instances from the treatment group $\overline{y}^{(1)}_i$. The
outputs of the proposed model are the predictions of the conditional outcomes
$\hat{Q}(0, \mathbf{x}_i, \overline{y}^{(0)}_i; \theta)$ and $\hat{Q}(1,
\mathbf{x}_i, \overline{y}^{(1)}_i; \theta)$ and the prediction of the
propensity score $\hat{g}( \mathbf{x}_i; \theta)$, where $\theta$ is the vector
with the network's parameters. These are computed from the model's three-head
architecture.

\begin{figure}[ht]
	\centering
	\includegraphics[width=10cm]{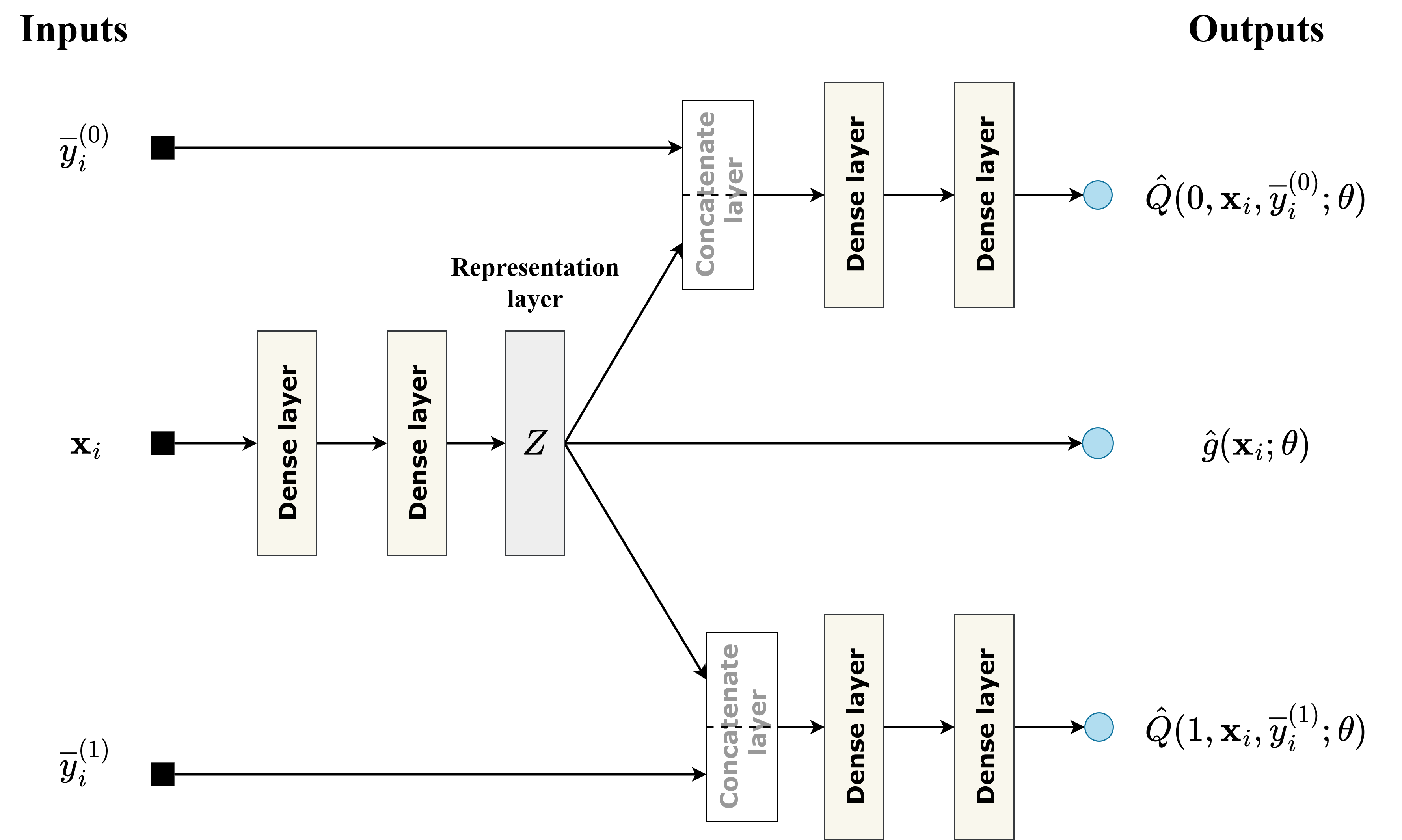}
	\caption{NN-Dragonnet architecture}\label{Fig:NN-Dragonnet}
\end{figure}

Figure \ref{Fig:NN-Dragonnet} illustrates the architecture of the proposed
NN-Dragonnet model. Initially, the instance $\mathbf{x}_i$ is processed by
three dense layers of 200 neurons each, with Exponential Linear Unit (ELU)
activation function for producing the representation layer $Z(\mathbf{x}_i) \in
\mathbb{R}^p$. Then, $Z(\mathbf{x}_i)$ is concatenated with
$\overline{y}^{(0)}_i$ and processed by two dense layers of 100 neurons with
ELU activation function and kernel regularizer of $10^{-2}$. Next, an output
layer of one neuron with linear activation provides the prediction of the
outcome $\hat{Q}(0, \mathbf{x}_i, \overline{y}^{(0)}_i; \theta)$. 

Similarly, $Z(\mathbf{x}_i)$ is concatenated with $\overline{y}^{(1)}$ and
processed by two dense layers of 100 neurons with ELU activation function and
kernel regularizer of $10^{-2}$ and a linear output for providing the
prediction of the outcome $\hat{Q}(1, \mathbf{x}_i, \overline{y}^{(1)}_i;
\theta)$. Furthermore, $Z(\mathbf{x}_i)$ is used for providing the prediction
for the propensity score $\hat{g}(\mathbf{x}_i; \theta)$, using linear
activation followed by a sigmoid.


The NN-Dragonnet model is trained using the following loss, which is a
modification of \textit{targeted regularization} \cite{shi2019adapting},
namely:
\begin{eqnarray}
	\hat{\theta},\hat{\epsilon} &=&
	\argmin_{\theta,\epsilon} \left[
	\hat{R}\left(\theta; \mathbf{X}, \overline{\mathbf{y}}^{(0)}, \overline{\mathbf{y}}^{(1)}\right)\right. \nonumber\\ 
	&+& \left. \beta\frac{1}{n} \sum_i
	\gamma\left(y_i, t_i, \mathbf{x}_{i}, {\overline{\mathbf{y}}_{i}}^{(0)},  {\overline{\mathbf{y}}_{i}}^{(1)}; \theta,\epsilon\right)\label{Eq:Regularized loss}
	\right]
\end{eqnarray}
where $\beta>0$ and $\epsilon > 0$ are hyper-parameters and
\begin{eqnarray}
	\hat{R}(\theta; \mathbf{X}, \overline{\mathbf{y}}^{(0)}, \overline{\mathbf{y}}^{(1)}) &=& \frac{1}{n} \sum_i \biggl[ \left( \hat{Q}\left(t_i,\mathbf{x}_{i}, {\overline{\mathbf{y}}_{i}}^{(0)}, {\overline{\mathbf{y}}_{i}}^{(1)};\theta\right) - y_i \right)^2\nonumber\\
	&+& \alpha f(\hat{g}(\mathbf{x}_{i};\theta), t_i) \biggl] 
\end{eqnarray}
and
\begin{eqnarray*}
	\gamma(y_i, t_i,\mathbf{x}_{i}, \overline{\mathbf{y}}^{(0)}, \overline{\mathbf{y}}^{(1)}; \theta,\epsilon) & = & \left(y_i - \tilde{Q}\left(t_i, \mathbf{x}_{i}, \overline{\mathbf{y}}^{(0)}, \overline{\mathbf{y}}^{(1)};\theta,\epsilon\right)\right)^2  \\[5pt]
	\tilde{Q}(t_i,\mathbf{x}_{i}, \overline{\mathbf{y}}^{(0)}, \overline{\mathbf{y}}^{(1)};\theta,\epsilon) &=& \hat{Q} \left(t_i, \mathbf{x}_{i}, \overline{\mathbf{y}}^{(0)}, \overline{\mathbf{y}}^{(1)};\theta\right)\\
	&+& \epsilon\left[ \frac{t_i}{\hat{g}(\mathbf{x}_{i};\theta)} -
	\frac{1-t_i}{1-\hat{g}(\mathbf{x}_{i};\theta)} \right]
\end{eqnarray*}

\subsubsection{NN-TARnet}

We present the NN-TARnet model, which is based on the adoption of NNCI
methodology in the TARnet model. The inputs are the instances $\mathbf{x}_i$,
the average outcomes of its $k$ neighboring instances from control group
$\overline{y}^{(0)}_i$ and the average outcomes of its $k$ neighboring
instances from treatment group $\overline{y}^{(1)}_i$; while the outputs are
the conditional outcomes for the two different groups of subjects, control
$\hat{Q}(0, \mathbf{x}_i, \overline{y}^{(0)}_i; \theta)$ and treatment
$\hat{Q}(1, \mathbf{x}_i, \overline{y}^{(1)}_i; \theta)$ groups, where $\theta$
is the vector with the network's parameters.

\begin{figure}[ht]
	\centering
	\includegraphics[width=10cm]{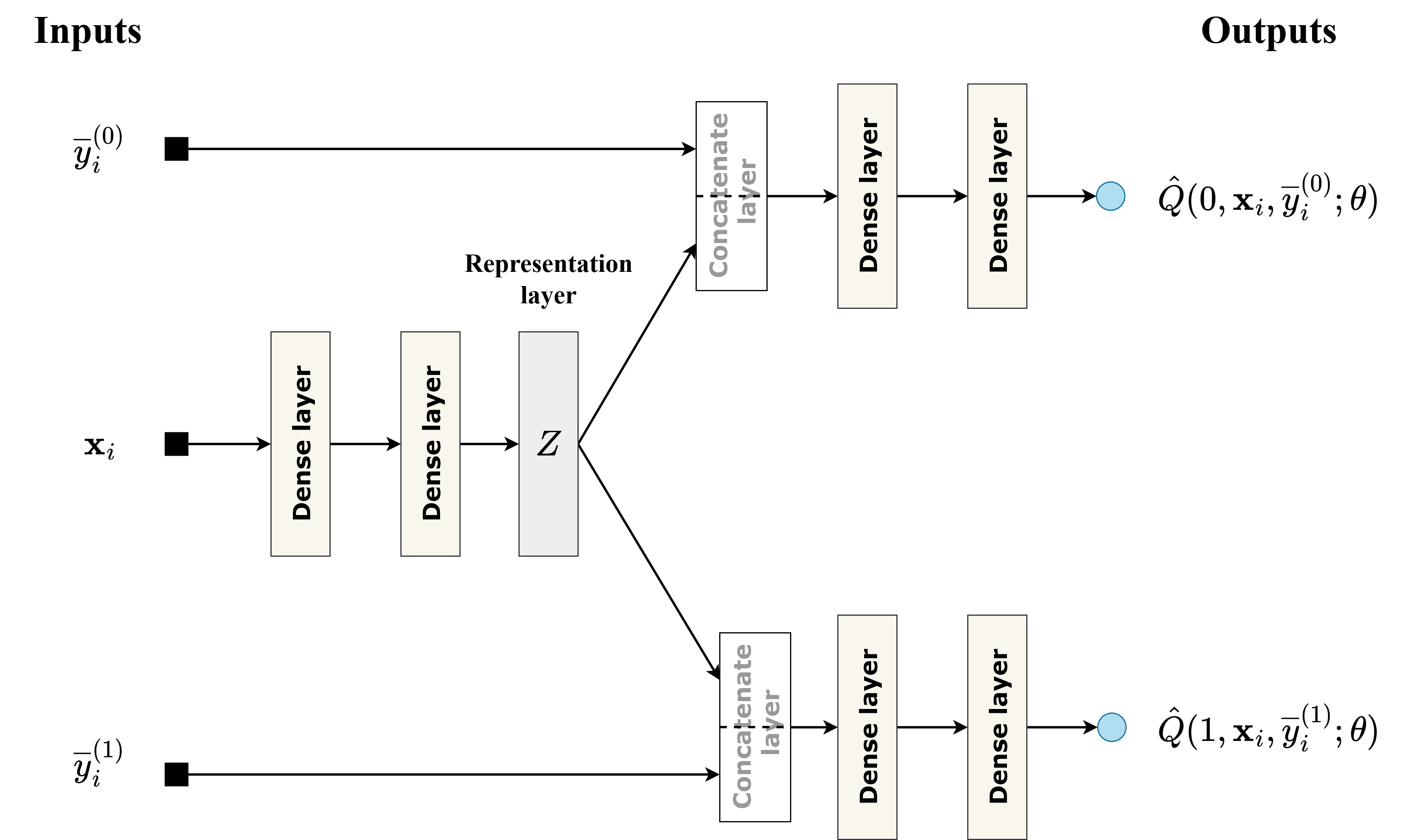}
	\caption{NN-TARnet architecture}\label{Fig:NN-TARnet}
\end{figure}

Figure \ref{Fig:NN-TARnet} presents the architecture of NN-TARnet neural
network model. As regards to NN-TARnet model, $\mathbf{x}_i$ is processed by
three dense layers of 200 neurons with ELU activation function and the
representation layer $Z(\mathbf{x}_i) \in \mathbb{R}^p$ is produced.

Next, $Z(\mathbf{x}_i)$ is concatenated with $\overline{y}^{(0)}_i$ and
processed by two dense layers of 100 neurons with ELU activation function and
kernel regularizer of $10^{-2}$. Next, an output layer of one neuron with
linear activation provides the prediction of the outcome $\hat{Q}(0,
\mathbf{x}_i, \overline{y}^{(0)}_i; \theta)$.
Likewise, $Z(\mathbf{x}_i)$ is concatenated with $\overline{y}^{(1)}$ and
processed by two dense layers of 100 neurons with ELU activation function and
kernel regularizer of $10^{-2}$ and a linear output for providing the
prediction of the outcome $\hat{Q}(1, \mathbf{x}_i, \overline{y}^{(1)}_i;
\theta)$. \\

Essentially, NN-TARnet uses the same architecture as NN-Dragonnet for the
estimation of conditional outcomes, with the difference that NN-Dragonnet uses
the propensity score for targeted regularization.

\subsubsection{NN-NEDnet}

Next, we present NN-NEDnet, a model which is based on the adoption of NNCI
methodology by NEDnet. NN-NEDnet shares the same architecture with
NN-Dragonnet, with the difference that it is not based on an end-to-end
philosophy, but on a multi-stage procedure.

Initially, NN-NEDnet is trained utilizing a treatment prediction task {(Figure
	4(a))}. In this case, the representation layer is trained using cross-entropy loss on the
propensity score $\hat{g}(\mathbf{x}_i\ ; \theta)$. Then, the conditional outcome head is removed and substituted by an outcome prediction network, similar with the one used by NN-Dragonnet {(Figure
	4(b))}. The difference is that the representation layers are frozen and
concatenated with the inputs ${\overline{y}_{i}}^{(0)}$ and
${\overline{y}_{i}}^{(1)}$ for estimating the conditional outcomes $\hat{Q}(0,
\mathbf{x}_i, \overline{y}^{(0)}_i; \theta)$ and $\hat{Q}(1, \mathbf{x}_i,
\overline{y}^{(1)}_i; \theta)$, respectively, where $\theta$ is the vectors of
NN-NEDnet's parameters. In this case, the representation
layer is frozen and the final model is frozen and trained using the mean
squared error loss on the factual outcomes, i.e.,
$$
\hat{\mathcal{L}}(\theta; \mathbf{X}, \overline{\mathbf{y}}^{(0)}, \overline{\mathbf{y}}^{(1)}) = \frac{1}{n} \sum_i \biggl[ \left( \hat{Q}\left(t_i,\mathbf{x}_{i}, {\overline{\mathbf{y}}_{i}}^{(0)}, {\overline{\mathbf{y}}_{i}}^{(1)};\theta\right) - y_i \right)^2
$$

\begin{figure}[ht]
	\centering
	\includegraphics[width=10cm]{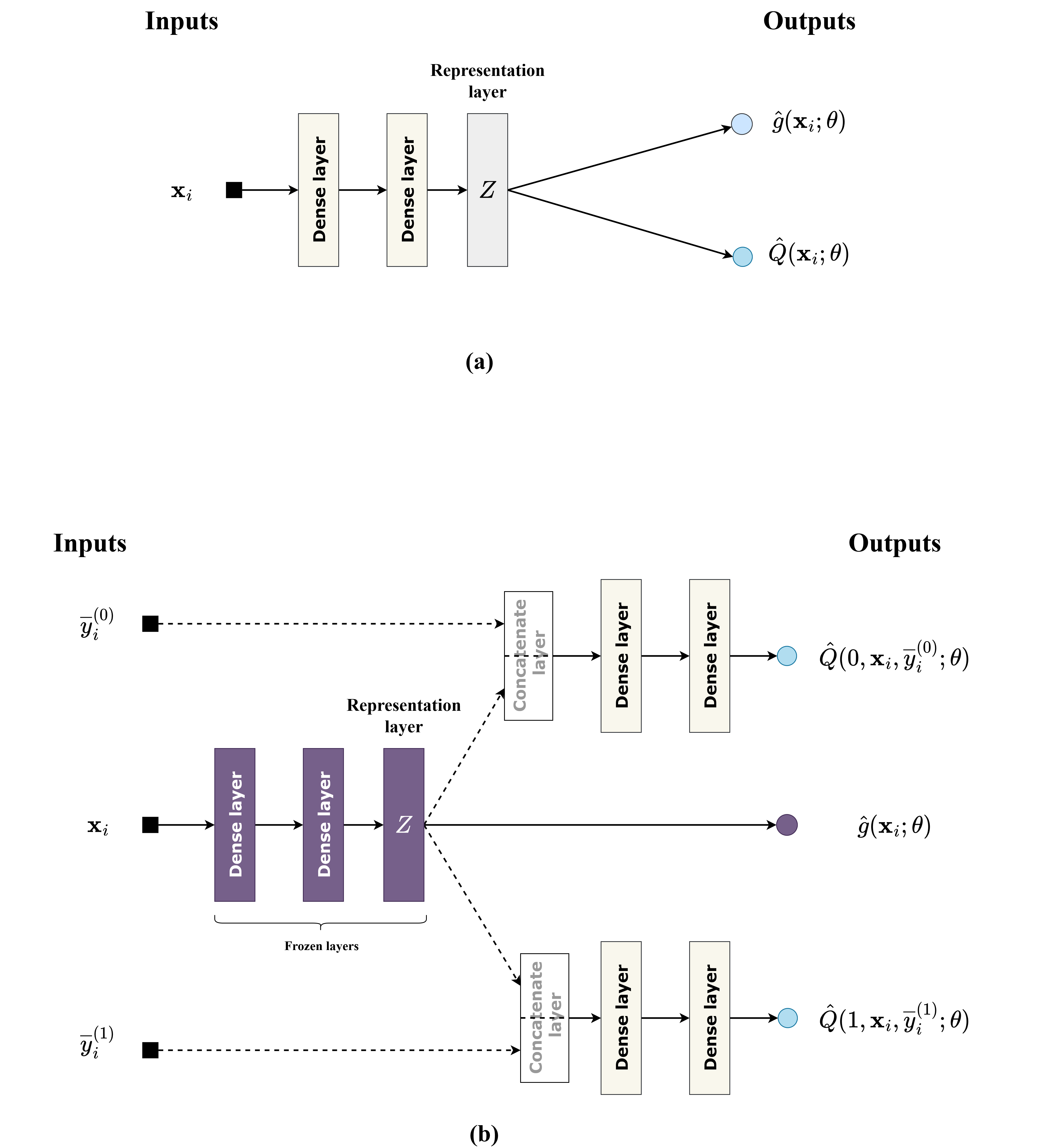}
	\caption{NN-NEDnet architecture. (a) Treatment prediction mask, used for
		training. (b) Outcome prediction, with frozen representation layers.}\label{Fig:NN-NedNet}
\end{figure}

\section{Evaluation Methodology \& Datasets}\label{sec:evaluation}

In this section, we comprehensively describe the tools for the evaluation
of causal inference models.

\subsection{Performance Profiles \& Statistical Analysis}
\label{sec:profiles}
The most commonly used performance metrics for evaluating the estimation of treatment
effects include the absolute error in the estimation of {ATE} and
the error of Precision in Estimation of Heterogeneous Effect {(PEHE)}, \cite{johansson2016learning,louizos2017causal,shalit2017estimating,yoon2018ganite,KiriakidouAIAI22} which are respectively defined as follows:
\begin{equation}
	\left\vert \epsilon_{ATE}\right\vert = \left\vert \frac{1}{n}\sum_{i=1}^{n}(y_{i}^{(1)} - y_{i}^{(0)}) -
	\frac{1}{n}\sum_{i=1}^{n}(\hat{y}_{i}^{(1)} - \hat{y}_{i}^{(0)}) \right\vert
\end{equation}
and 
\begin{equation}
	\epsilon_{PEHE} = \displaystyle \frac{1}{n} \sum_{i=1}^{n} \biggl(
	(y_{i}^{(1)} - y_{i}^{(0)}) -  (\hat{y}_{i}^{(1)} - \hat{y}_{i}^{(0)})\biggl)^{2}
\end{equation}
where $\hat{y}_{i}^{(t)}$ indicates the model's outcome prediction for the
$i$-th sample for treatment $t$. {Notice that $\epsilon_{ATE}$ and $\epsilon_{PEHE}$ concern the evaluation of the estimation of average (ATE) and individual (ITE) treatment effects, respectively.}

In the literature, it is common to use ensembles of experiments for the evaluation
of the models' effectiveness which is usually carried out by calculating the
average $\vert \epsilon_{ATE}\vert $ and $\epsilon_{PEHE}$ across the
experiments. Nonetheless, this approach may be misleading in some cases,
since all problems are equally considered for the models' evaluation
and a small number of them tend to dominate the results \cite{kiriakidou2022evaluation,KiriakidouAIAI22}.

To address this problem, we adopt the evaluation framework for causal inference models
proposed in  \cite{kiriakidou2022evaluation}, using the performance profiles of Dolan and
Mor{\'e} \cite{dolan2002benchmarking} and a comprehensive statistical analysis based on nonparametric and post-hoc tests. 
The performance profiles provide us information such as probability of success, effectiveness and robustness in compact form
\cite{livieris2018improving,livieris2020advanced}. Each profile plots the fraction $P$ of experiments
for which any given model is within a factor $\tau$ of the best model.
%
%
Statistical analysis is conducted to examine the hypothesis that a
pair of models perform equally well and provide statistical evidence about the
superiority of the proposed model \cite{fernandez2017pareto,livieris2019employing,livieris2020improved,vuttipittayamongkol2020improved,tampakas2019prediction}. 
Firstly, we apply the non-parametric Friedman Aligned-Ranks (FAR) test \cite{hodges1962rank} to rank the evaluated models from the best to the worst performance
and the Finner post-hoc test \cite{finner1993monotonicity} to examine whether
there are statistically significant differences among the evaluated models'
performance.

\subsection{Datasets}


The fundamental problem of causal inference is that in practice, we only have
access to the factual outcomes, which prohibits model evaluation. To address
this issue, we used simulated or semi-synthetic datasets, which include both
outcomes for each sample. 
{It is worth mentioning that the selected datasets are
	the most widely used in the field of causal inference and have been chosen by
	various researchers \cite{yao2018representation,shi2019adapting,johansson2016learning,shalit2017estimating,louizos2017causal}
	for the evaluation of their proposed models.} 

\textbf{IHDP dataset}. It constitutes a collection of semi-simulated datasets, which was
constructed from the Infant Health and Development Program
\cite{hill2011bayesian}.  Each dataset is composed by 608 units in the control
group and 139 units in the treatment group, while the 25 covariates are
collected from a real-world randomized experiment.  Furthermore, the effect of
home visits by specialists was studied on future cognitive test scores.  In our
experiments, we used the setting ``A'' in NPCI package \cite{dorie2016npci}
composed by 1000 realizations. Notice that 80\% of the data were used for training the models, while the rest 20\% was used for testing.

\textbf{Synthetic dataset}. It consists of a collection of toy datasets, which was
originally introduced by Louizos et al. \cite{louizos2017causal}.  Its
generation is based on the hidden confounder variable $W$ using the following
process:
\begin{eqnarray*}
	w_i &\sim& \text{Bern}(0.5) \\ 
	t_i\, |\, w_i &\sim& \text{Bern}(0.75 w_i + 0.25(1-w_i)) \\
	x_i\, |\, w_i &\sim& \mathcal{N}(w_i, \sigma_{z_1}^2 w_i + \sigma_{z_0}^2 (1-w_i)) \\ 
	y_i\, |\, t_i, w_i &\sim& \text{Bern}(\text{Sigmoid}(3(w_i+2(2t_i-1)))
\end{eqnarray*}
,where Sigmoid is the logistic sigmoid function and $\sigma_{z_0} = 3$, $\sigma_{z_1}= 5$. Notice that the treatment variable $T$ and the proxy to the confounder $X$ constitute a mixture of Bernoulli and Gaussian distribution, respectively. Finally, 90\% of the data were used for training the models, while the rest 10\% was used for testing.

\textbf{ACIC dataset}. This dataset was developed for the \textit{2018 Atlantic
	Causal Inference Conference competition data}
\cite{shimoni2018benchmarking}. ACIC is a collection of semi-synthetic
datasets, which were received from the linked birth and infant death data
\cite{macdorman1998infant}.  Every competition dataset is a sample from a
distinct distribution and data are generated through a generating process, by
possessing different treatment selection and outcome functions.  Following Shi
et al. \cite{shi2019adapting} for each data generating process setting, four
and eight datasets were randomly picked of size 5000 and 10000, respectively. Notice that 80\% of the data were used for training the models, while the rest 20\% was used for testing.

\section{Numerical Experiments}\label{sec:experiments}

In this section, we evaluate the prediction performance of the neural network
models, which integrate the proposed NNCI methodology against the corresponding
baseline models, on IHDP, Synthetic and ACIC datasets. More specifically, we
evaluate the performance of the proposed NN-Dragonnet, NN-TARnet, NN-NEDnet
against Dragonnet, TARnet and NEDnet, respectively.

{The implementation code was written in Python 3.7 using Tensorflow Keras library 
	\cite{gulli2017deep} and run on a PC (3.2GHz Quad-Core processor, 16GB RAM)
	using Windows operating system.} Notice that Dragonnet, TARnet and NEDNet were used with their default optimized
parameter settings, while for the proposed models the number of neighbors was
set to $k=11$ and targeted regularization was implemented with $\alpha = 1$ and
$\beta = 1$ \cite{shalit2017estimating}.
%

The curves in the following figures have the following meaning
\begin{itemize}
	\item ``{Dragonnet}'' stands for the state-of-the-art neural
	network-based model Dragonnet \cite{shi2019adapting}.
	
	\item ``{TARnet}'' stands for the neural network-based model TARnet \cite{shalit2017estimating}. 
	
	\item ``{NEDnet}'' stands for the two-stage neural
	network-based model NEDnet \cite{shi2019adapting}. 
	
	\item ``{NN-Dragonnet}'' stands for the proposed NN-Dragonnet
	model, which modifies Dragonnet using the NNCI methodology. 
	
	\item ``{NN-TARnet}'' stands for the NN-TARnet model, which
	modifies TARnet using the NNCI methodology. 
	
	\item ``{NN-NEDnet}'' stands for the NN-NEDnet model, which
	modifies NEDnet using the NNCI methodology.
\end{itemize}
Additionally, it is worth mentioning that the proposed models were implemented
using three different distance metrics i.e Euclidean, Manhattan and Chebychev { as in \cite{KiriakidouAIAI22}. These distances belong to the class of Minkowski distances 
	and constitute the most widely used distances proposed in the literature \cite{pandit2011comparative,singh2013k}.}

\subsection{{Evaluation} on IHDP dataset}

\begin{figure*}[t]
	\subfigure[$\ \vert \epsilon_{ATE} \vert$]{\includegraphics[width=0.5\linewidth]{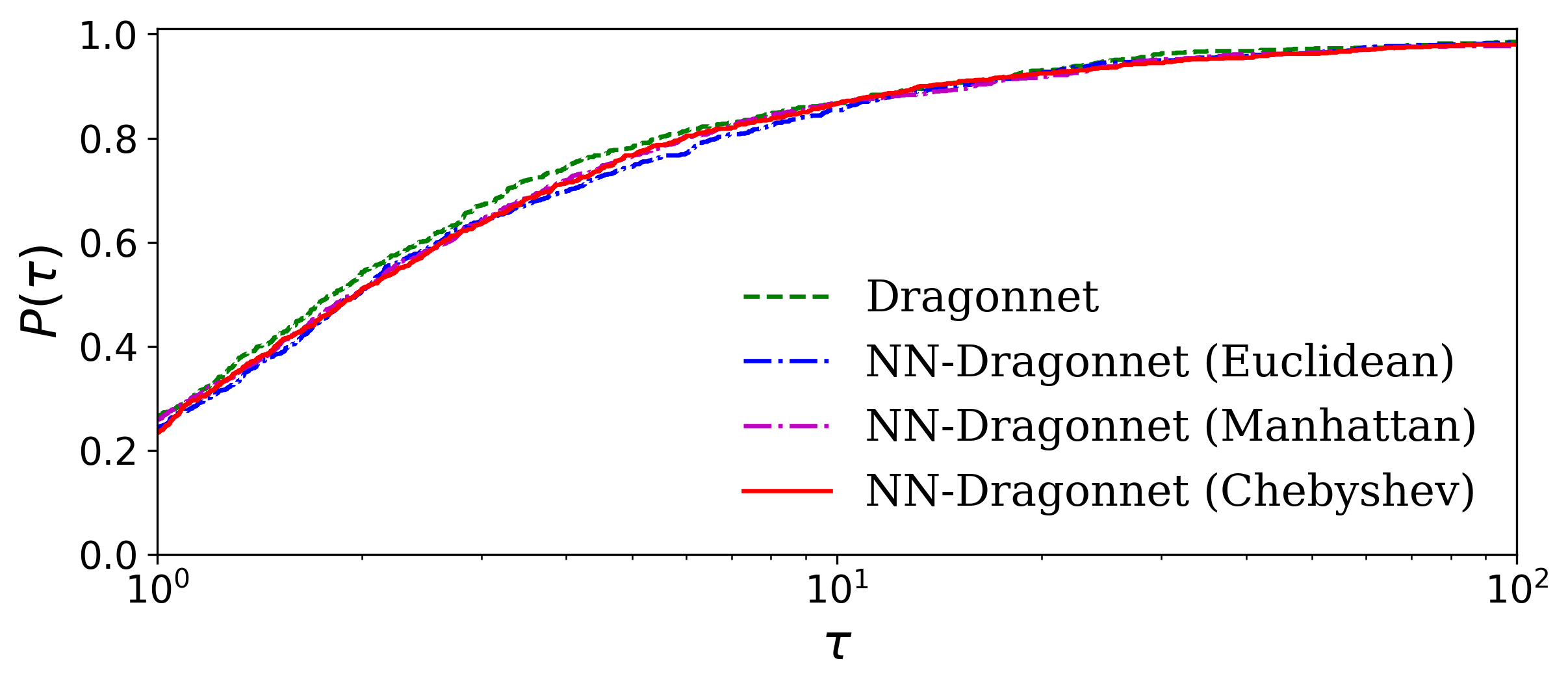}}\hfill
	\subfigure[$\ \epsilon_{PEHE} $]{\includegraphics[width=0.5\linewidth]{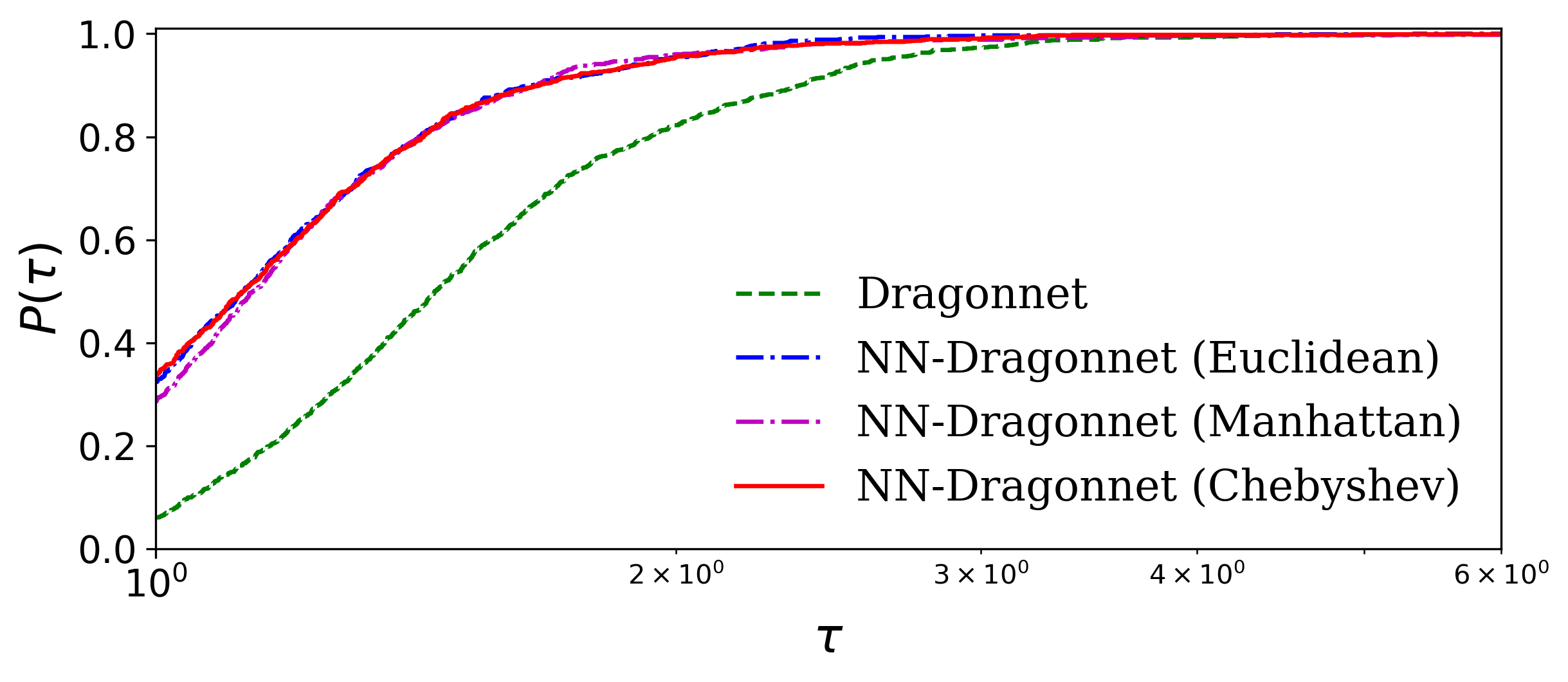}}
	\subfigure[$\ \vert \epsilon_{ATE} \vert$]{\includegraphics[width=0.48\linewidth]{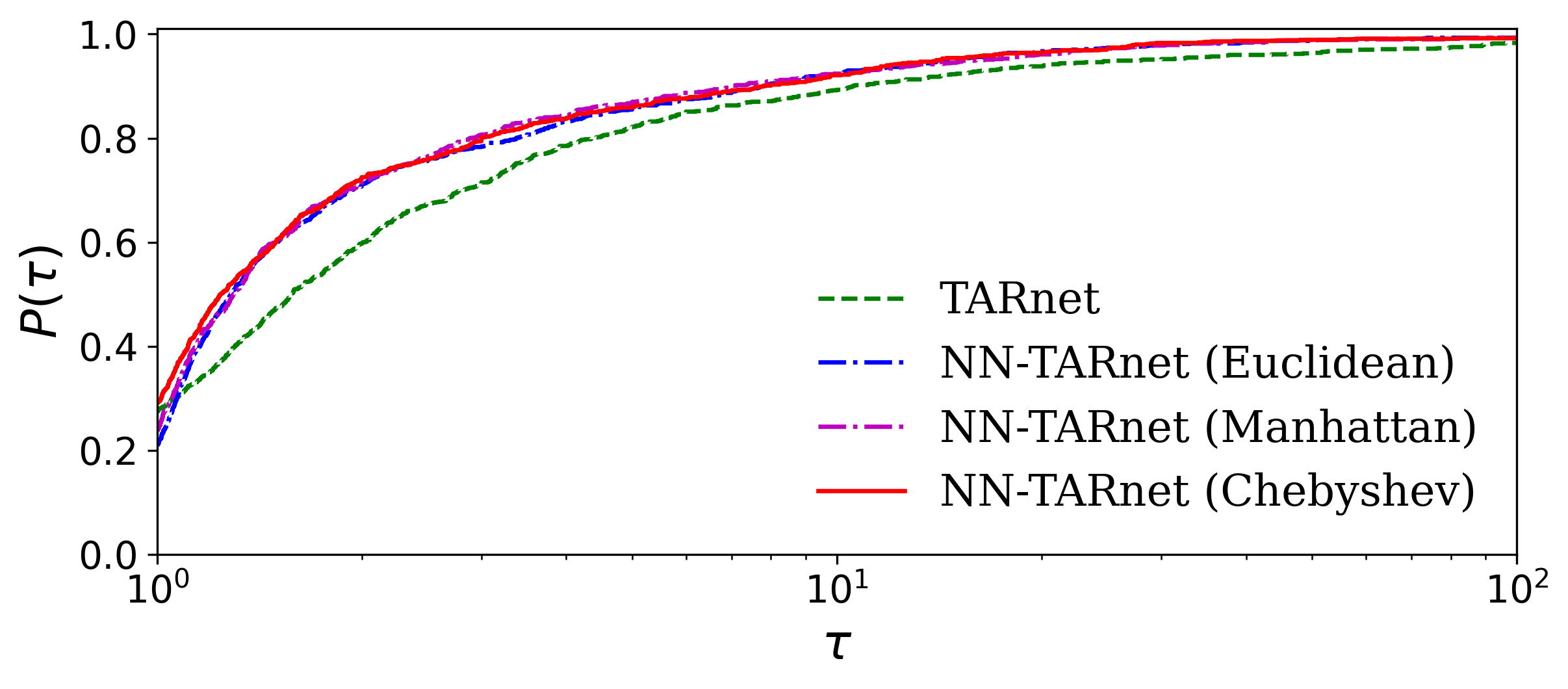}}\hfill
	\subfigure[$\ \epsilon_{PEHE} $]{\includegraphics[width=0.48\linewidth]{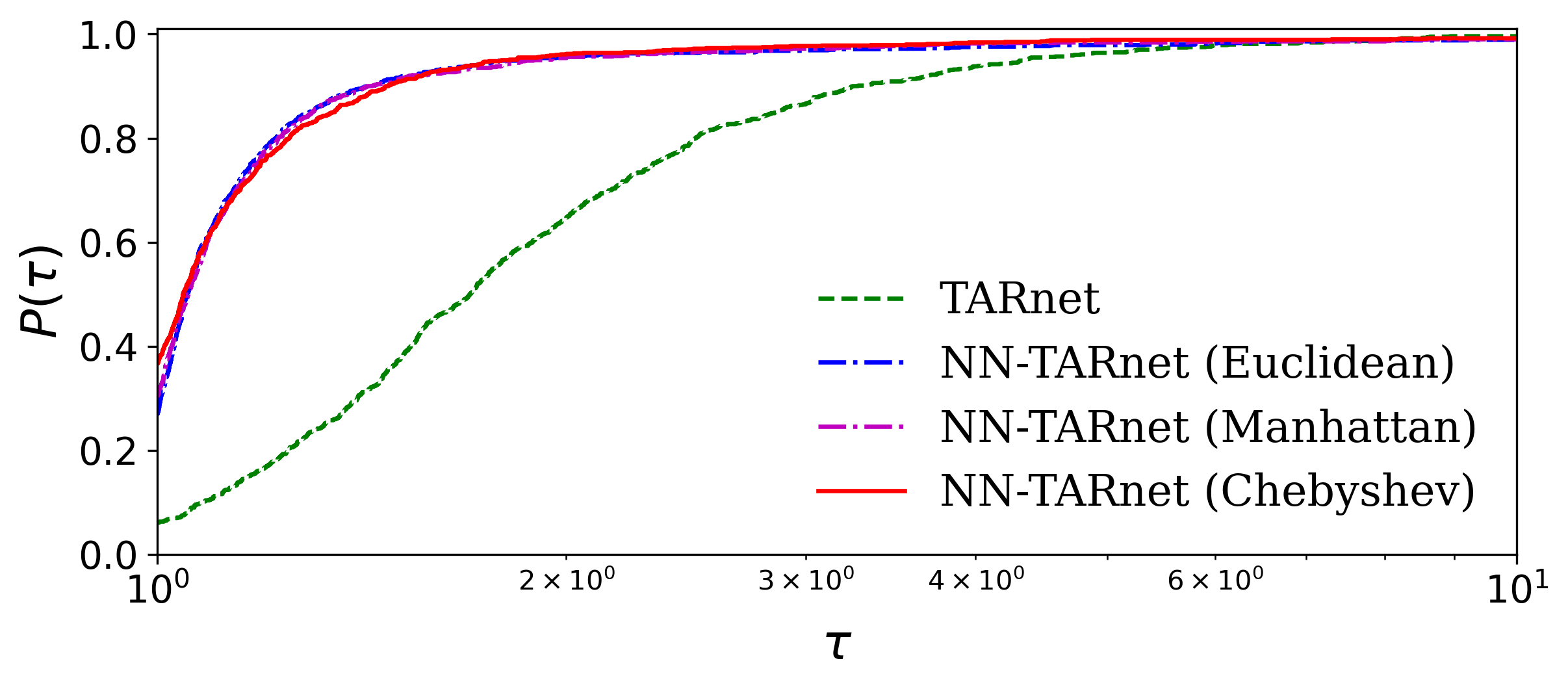}}
	\subfigure[$\ \vert \epsilon_{ATE} \vert$]{\includegraphics[width=0.48\linewidth]{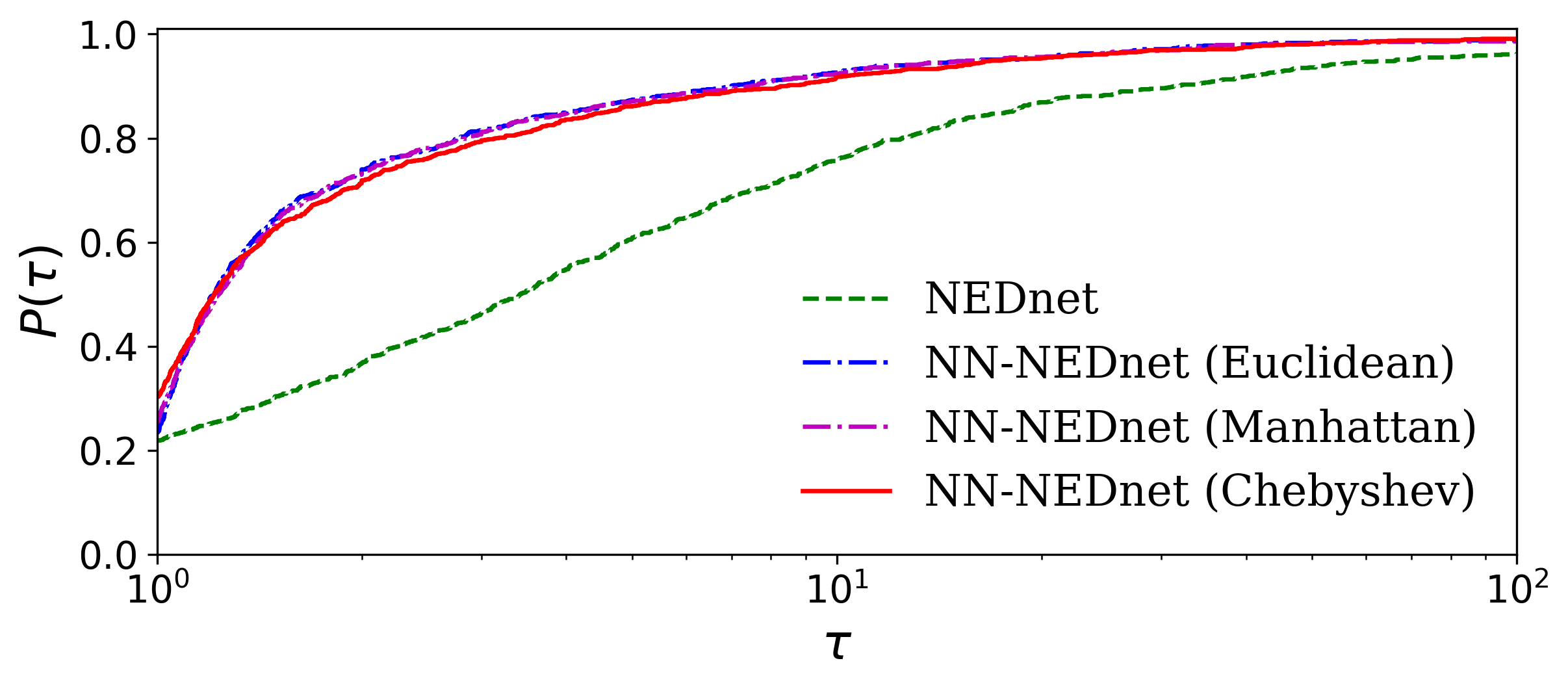}}\hfill
	\subfigure[$\ \epsilon_{PEHE} $]{\includegraphics[width=0.48\linewidth]{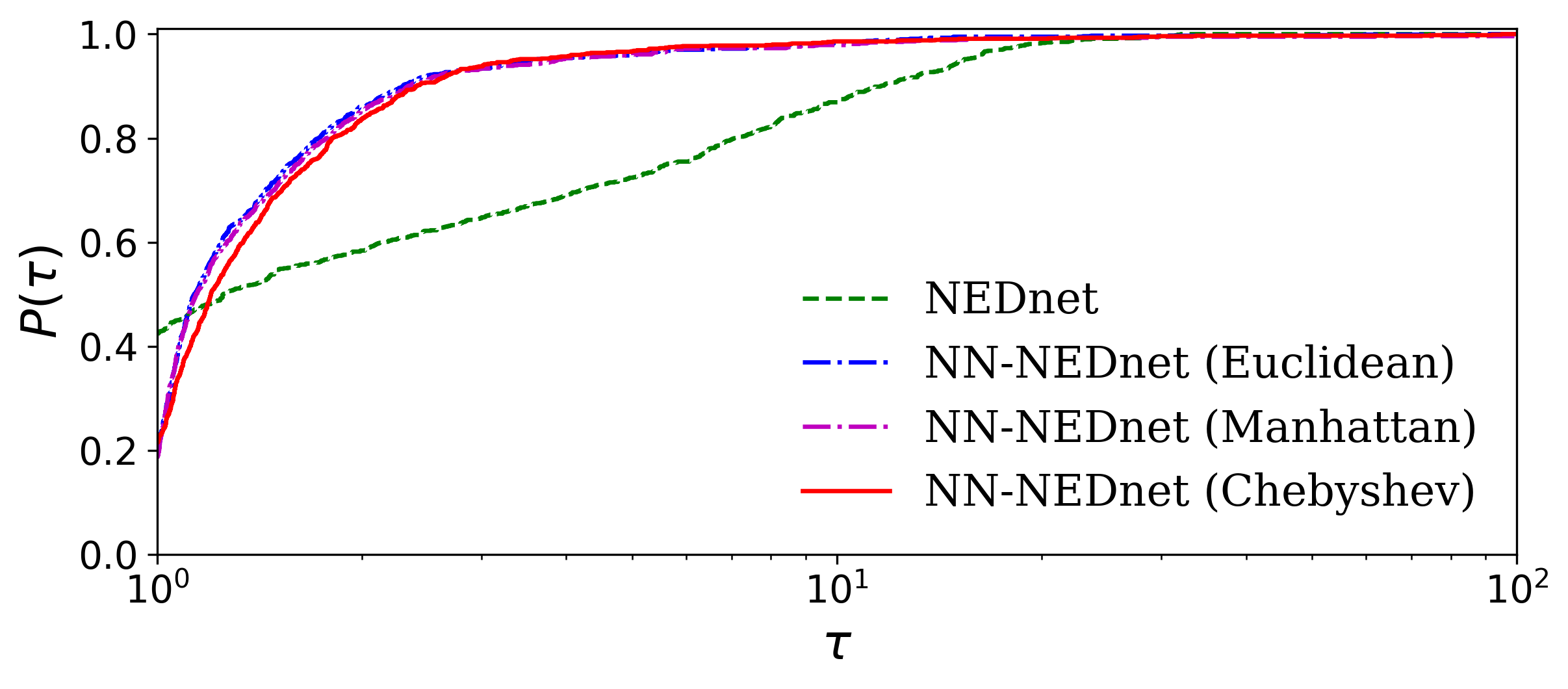}}
	\caption{Log$_{10}$ Performance profiles of models, based on $\vert \epsilon_{ATE} \vert $ and $\epsilon_{PEHE}$ for IHDP dataset} \label{fig:IHDP}
\end{figure*}

	%

{Figure \ref{fig:IHDP} 
	presents the performance profiles of NN-Dragonnet, NN-TARnet, NN-NEDnet and their corresponding
	baseline models Dragonnet, TARnet and NEDnet, based on the performance metrics
	$ \vert \epsilon_{ATE} \vert$ and $\epsilon_{PEHE}$. 
	All versions of NN-Dragonnet exhibit similar performance with state-of-the-art Dragonnet in terms of $ \vert \epsilon_{ATE} \vert$, while they considerably outperform Dragonnet in terms of  $\epsilon_{PEHE}$.
	Additionally, the adoption of NNCI methodology considerably improved the performance 
	of TARnet model, relative to both performance metrics since the curves of all versions of NN-TARnet lie on the top. NN-NEDnet solves almost the same percentage of benchmarks 
	with the best (lowest) $\vert \epsilon_{ATE} \vert$ for any used distance metric 
	and outperforms the baseline model NN-NEDnet model in terms of robustness. 
	Finally, it is worth mentioning that all proposed models exhibited similar performance using any distance metric.
	In detail all versions of NN-NEDnet outperform the baseline model since their curves lie on the top.}

{
	Tables \ref{Table:IHDP1} and \ref{Table:IHDP2} present the statistical analysis of the proposed causal inference models and their corresponding baseline models for IHDP dataset, in terms of 
	$ \vert \epsilon_{ATE} \vert$ and $\epsilon_{PEHE}$, respectively.
	As regards $\vert \epsilon_{ATE}\vert$, Dragonnet model reports the best rank according to FAR test. However, Finner post-hoc test suggests that there are no statistically significant differences in the performance between Dragonnet and all versions of NN-Dragonnet model, which implies that all models perform similarly. NN-TARnet and NN-NEDnet report the highest probability-based ranking,
	outperforming TARnet and NEDnet, respectively. Additionally, Finner post-hoc test suggests that there are statistically significant differences in the performance between all versions of both NN-TARnet and NN-NEDnet with their corresponding baseline models.
	As regards $\epsilon_{PEHE}$ performance metric, the interpretation of Table~\ref{Table:IHDP2} reveals that NN-Dragonnet, NN-TARnet and NN-NEDnet are the top ranking models, outperforming the baseline models, Dragonnet, TARnet and NEDnet, respectively. Therefore, we able to conclude that we obtained strong statistical evidence that the adoption of the NNCI methodology considerably improved the performance of all baseline causal inference models.}

\clearpage

\begin{table}
	\centering
	\setlength{\tabcolsep}{3pt}
	\renewcommand{\arraystretch}{1}
	\caption{FAR test and Finner post-hoc test based on $\vert \epsilon_{ATE} \vert $ for IHDP dataset\hfill\label{Table:IHDP1}}
	{
		\begin{tabular}{lccc}
			\toprule
			\multirow{2}{*}{Model} &  \multirow{2}{*}{FAR} & \multicolumn{2}{c}{Finner post-hoc test}\\
			\cmidrule{3-4}                 &                       & $p_F$-value & $H_0$  \\
			\midrule
			{Dragonnet}                   & 1840.19   &      -      &     -           \\
			{NN-Dragonnet (Manhattan) }   & 1894.39   & 0.282160    &  Fail to reject \\
			{NN-Dragonnet (Chebyshev)}    & 1935.42   & 0.094191    &  Fail to reject \\
			{NN-Dragonnet (Euclidean)}    & 1947.98   & 0.094191    &  Fail to reject \\
			\midrule
			{NN-TARnet (Manhattan)}       & 1827.99     &      -      &     -           \\
			{NN-TARnet (Chebyshev)}       & 1831.38     & 0.947642    &  Fail to reject \\
			{NN-TARnet (Euclidean)}       & 1872.75     & 0.519053    &  Fail to reject \\
			{TARnet}                      & 2469.86     & 0.000000    &  Reject \\
			\midrule
			{NN-NEDnet (Manhattan)}       & 1839.56     &      -      &     -           \\
			{NN-NEDnet (Chebyshev)}       & 1845.91     & 0.938142    &  Fail to reject \\
			{NN-NEDnet (Euclidean)}       & 1849.75     & 0.938142    &  Fail to reject \\
			{NEDnet}                      & 2457.28     & 0.000000    &  Reject \\
			\bottomrule
		\end{tabular}
	}
\end{table}

\begin{table}
	\centering
	\setlength{\tabcolsep}{3pt}
	\renewcommand{\arraystretch}{1}
	\caption{FAR test and Finner post-hoc test based on  $\epsilon_{PEHE}$ for IHDP dataset\hfill\label{Table:IHDP2}}
	{
		\begin{tabular}{lccc}
			\toprule
			\multirow{2}{*}{Model} &  \multirow{2}{*}{FAR} & \multicolumn{2}{c}{Finner post-hoc test}\\
			\cmidrule{3-4}                 &                       & $p_F$-value & $H_0$  \\
			\midrule
			{NN-Dragonnet (Euclidean)}     & 1628.32          &      -      &     -           \\
			{NN-Dragonnet (Manhattan) }    & 1653.07          & 0.623290    &  Fail to reject \\
			{NN-Dragonnet (Chebyshev)}     & 1670.39          & 0.539643    &  Fail to reject \\
			{Dragonnet}                    & 2666.19          & 0.000000    &   Reject \\
			\midrule
			{NN-TARnet (Euclidean)}        & 1609.48         &      -      &     -           \\
			{NN-TARnet (Manhattan) }       & 1624.02         & 0.86625     &  Fail to reject \\
			{NN-TARnet (Chebyshev)}        & 1626.72         & 0.86625     &  Fail to reject \\
			{TARnet}                       & 3141.77         & 0.000000    &   Reject \\                     
			\midrule
			{NN-NEDnet (Euclidean)}        & 1908.98         &      -      &     -           \\
			{NN-NEDnet (Manhattan) }       & 1937.97         & 0.574488    &  Fail to reject \\
			{NN-NEDnet (Chebyshev)}        & 2001.56         & 0.107503    &  Fail to reject \\
			{NEDnet}                       & 2150.38         & 0.000000    &   Reject \\                   
			\bottomrule
		\end{tabular}
	}
	
\end{table}		

\subsection{{Evaluation} on Synthetic dataset}

\begin{figure*}[t]
	\subfigure[$\ \vert \epsilon_{ATE} \vert$]{\includegraphics[width=0.48\linewidth]{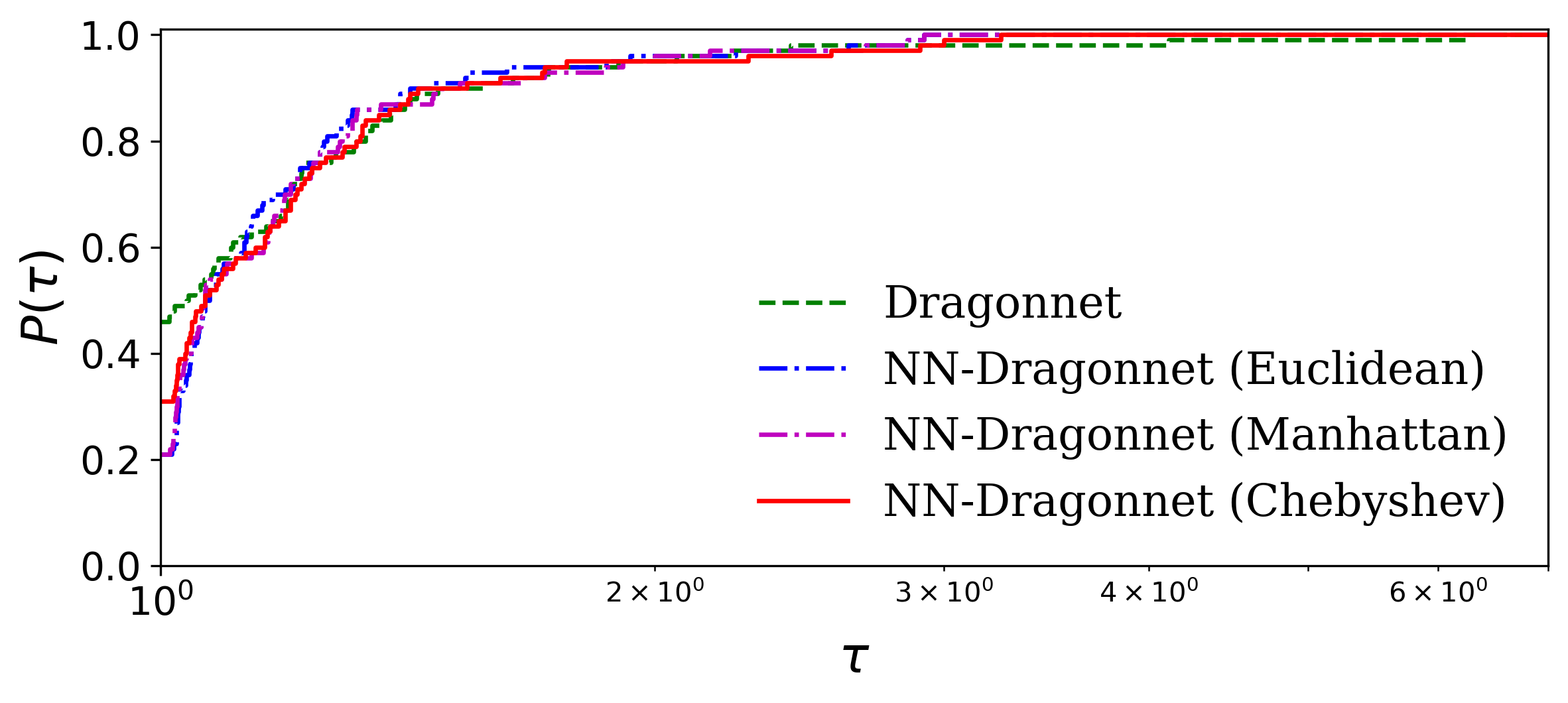}}\hfill
	\subfigure[$\ \epsilon_{PEHE} $]{\includegraphics[width=0.48\linewidth]{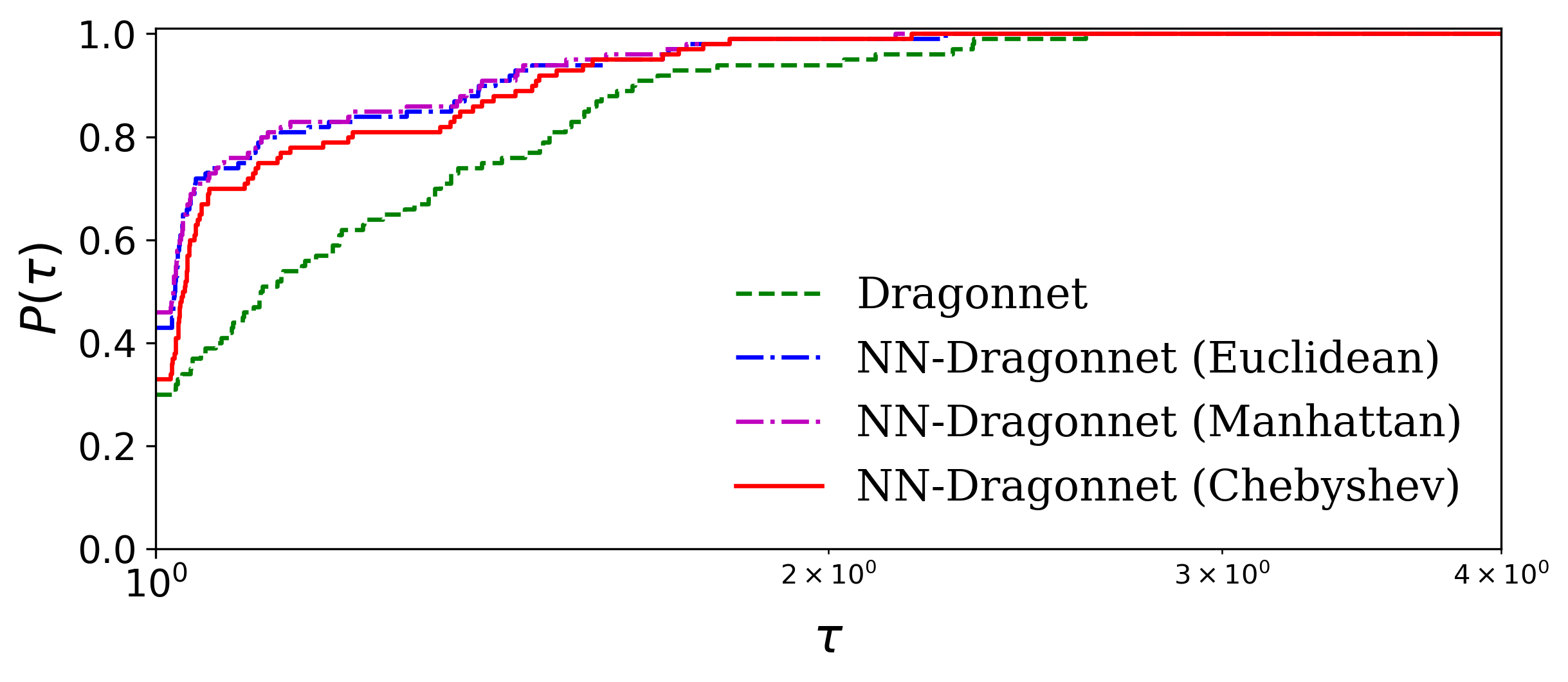}}
	\subfigure[$\ \vert \epsilon_{ATE} \vert$]{\includegraphics[width=0.48\linewidth]{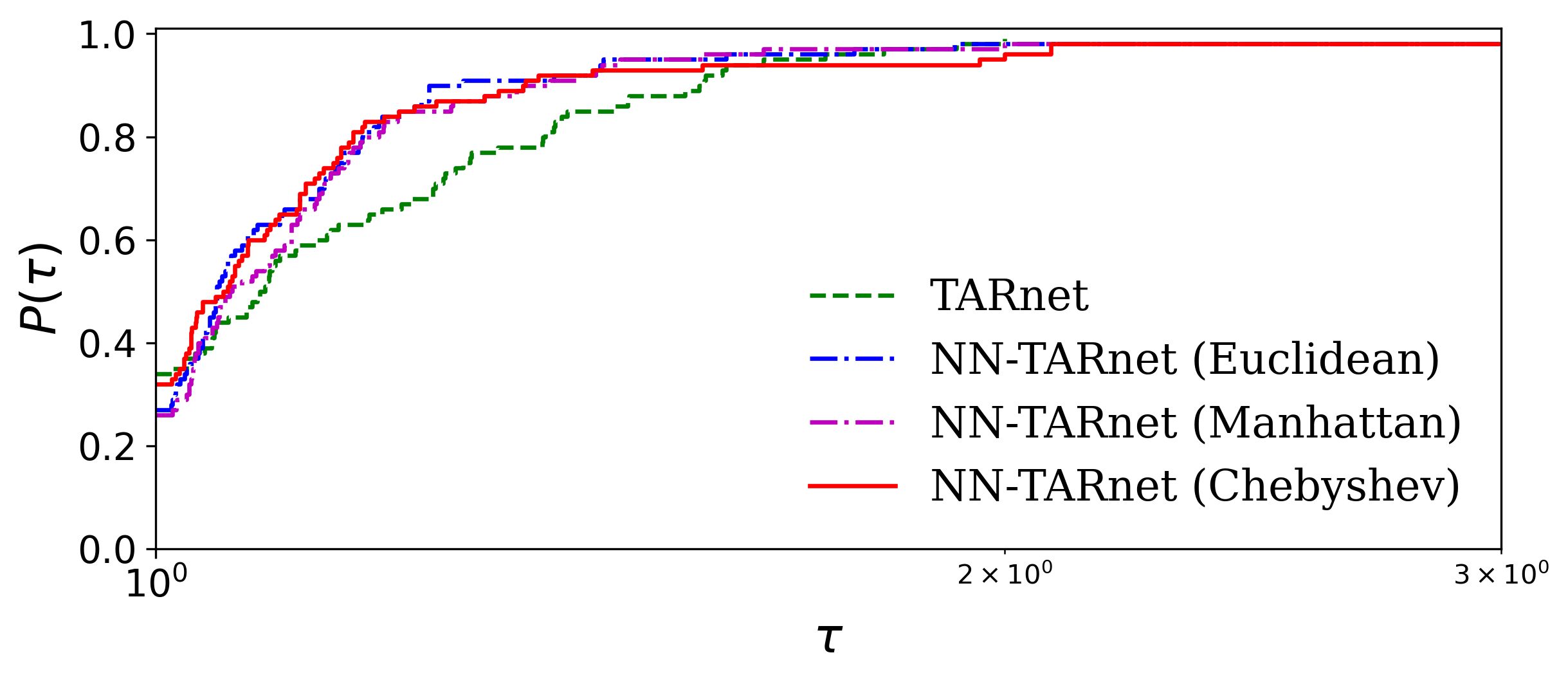}}\hfill
	\subfigure[$\ \epsilon_{PEHE} $]{\includegraphics[width=0.48\linewidth]{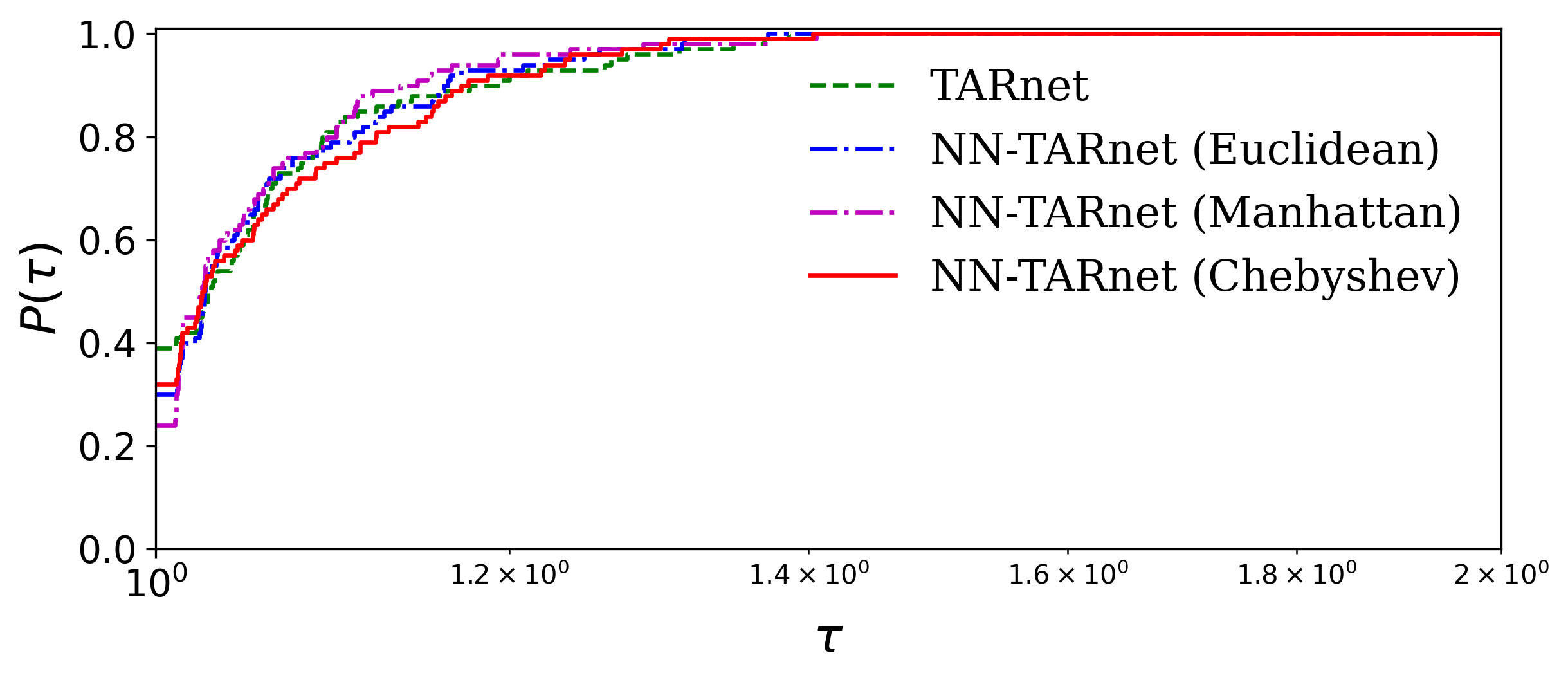}}
	\subfigure[$\ \vert \epsilon_{ATE} \vert$]{\includegraphics[width=0.48\linewidth]{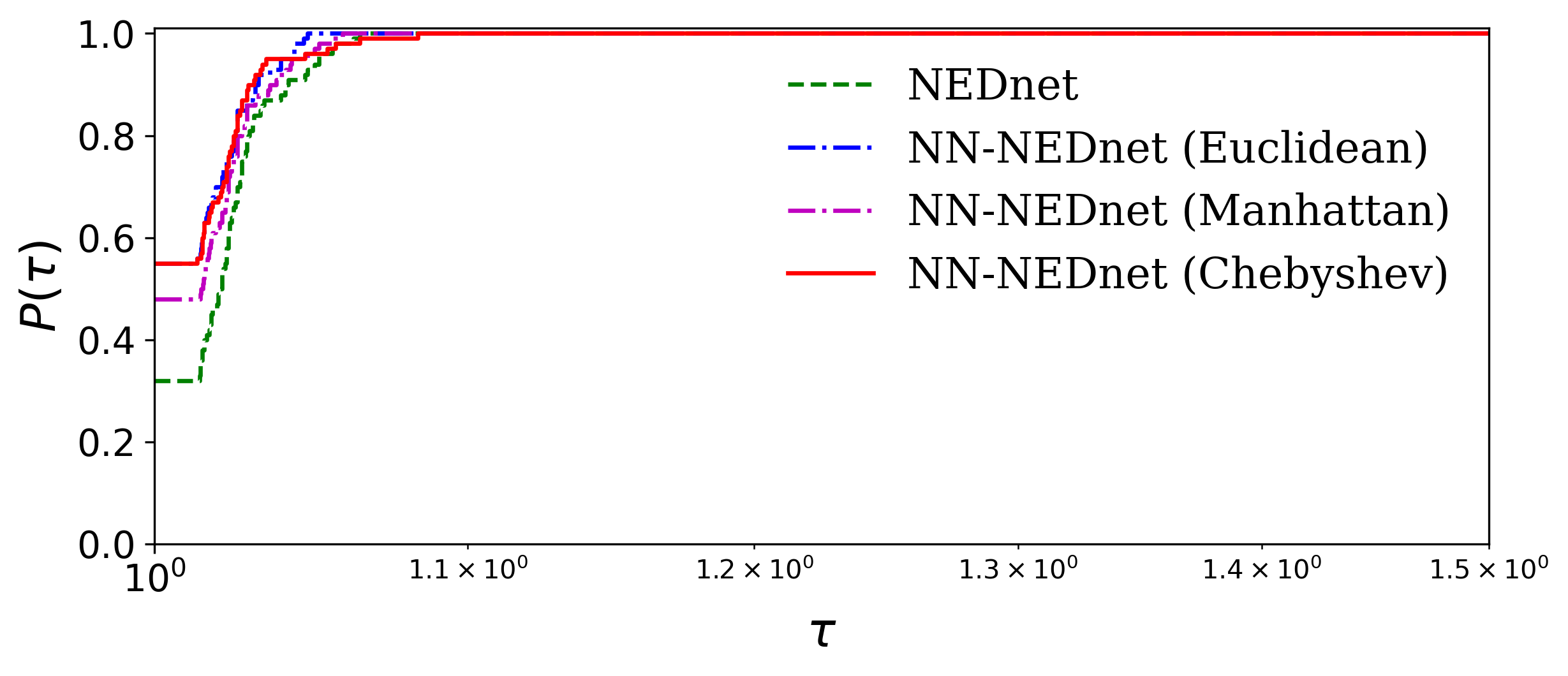}}\hfill
	\subfigure[$\ \epsilon_{PEHE} $]{\includegraphics[width=0.48\linewidth]{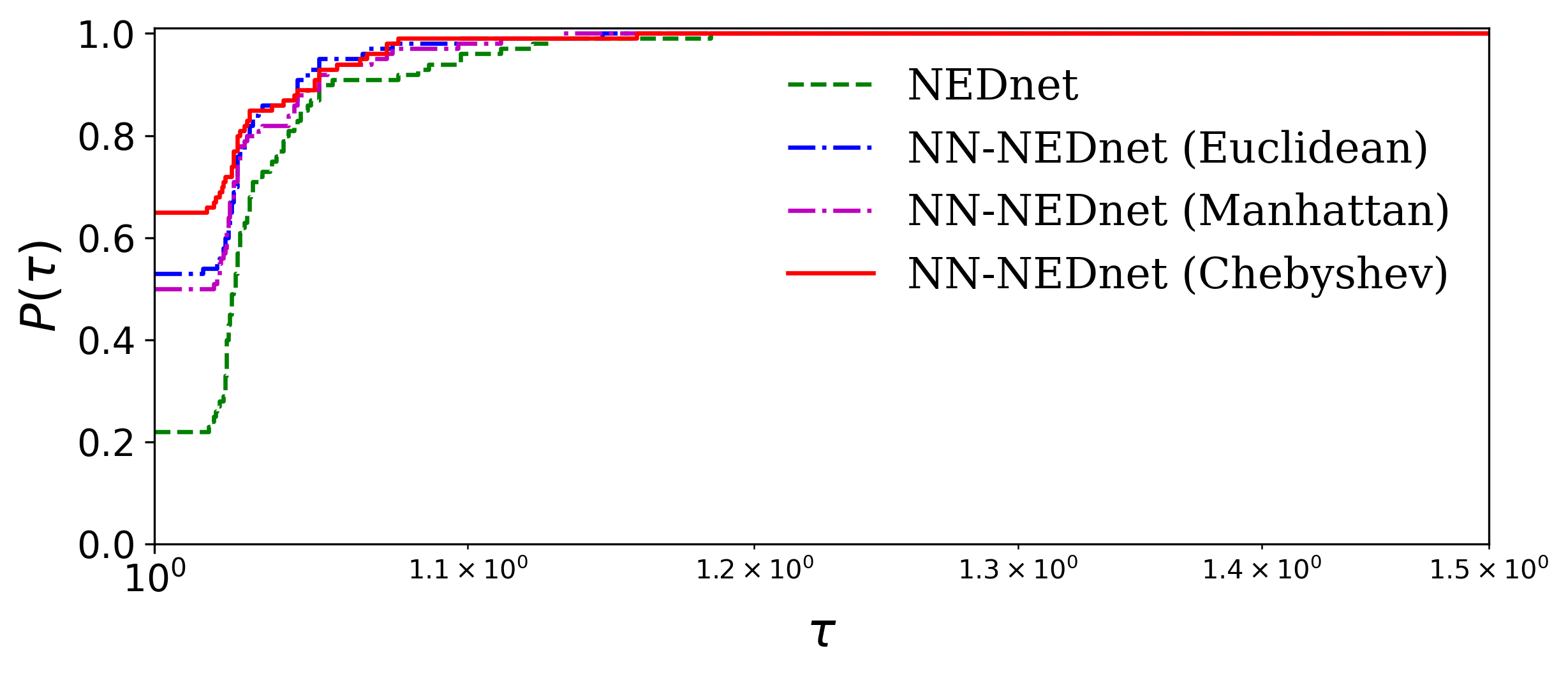}}
	\caption{Log$_{10}$ Performance profiles of models, based on $\vert \epsilon_{ATE} \vert $ and $\epsilon_{PEHE}$ for Synthetic dataset} \label{fig:Synthetic}
\end{figure*}

{Figure \ref{fig:Synthetic}
	demonstrates the performance profiles of the proposed NN-Dragonnet, NN-TARnet, NN-NEDnet and their corresponding
	state-of-the-art models Dragonnet, TARnet and NEDnet, in terms of 
	$ \vert \epsilon_{ATE} \vert$ and $\epsilon_{PEHE}$ performance metrics.
	Firstly, relative to $\vert \epsilon_{ATE} \vert $ all versions of the proposed models report
	almost identical performance while the use of Euclidean distance presents slightly better performance. NN-Dragonnet's versions exhibited similar performance with Dragonnet while
	NN-TARnet's and NN-NEDnet's versions slightly outperform the traditional TARnet and NEDnet, respectively in terms of efficiency. 
	In case of $\epsilon_{PEHE}$, NN-Dragonnet and NN-NEDnet outperformed Dragonnet and NEDnet, respectively using any distance metric since their curves lie on the top. 
	In detail, all versions of both NN-Dragonnet and NN-NEDnet presented the highest percentage of simulations with the best (lowest) $\epsilon_{PEHE}$. In contrast, NN-TARnet and TARnet reported almost identical performance, independent of the utilized distance metric.}

{Tables~\ref{Table:Synthetic1} presents the statistical analysis for the proposed causal inference models and their corresponding baseline models for Synthetic dataset, in terms of $ \vert \epsilon_{ATE} \vert$.
	NN-Dragonnet (Manhattan) is the top ranking model since it presents the best (lowest) FAR score.
	Nevertheless, Finner post-hoc tests suggests that there is no statistically significant differences in the performance of all evaluated models.
	NN-TARnet and NN-NEDnet present the highest FAR ranking using the Euclidean distance. 
	Additionally, Finner post-hoc test suggests that there exist significant differences between the performance of the proposed NN-TARnet and NN-NEDnet and the corresponding baseline models TARnet and NEDnet, independent of the utilized distance metrics. This implies that the adoption of the proposed NNCI methodology benefit the TARnet and NEDnet in terms of $ \vert \epsilon_{ATE} \vert$.}

\begin{table}
	\centering
	\setlength{\tabcolsep}{3pt}
	\renewcommand{\arraystretch}{1}
	\caption{FAR test and Finner post-hoc test based on $\vert \epsilon_{ATE} \vert $ for Synthetic dataset\hfill\label{Table:Synthetic1}}
	{
		\begin{tabular}{lccc}
			\toprule
			\multirow{2}{*}{Model} &  \multirow{2}{*}{FAR} & \multicolumn{2}{c}{Finner post-hoc test}\\
			\cmidrule{3-4}                 &                       & $p_F$-value & $H_0$  \\
			\midrule
			{NN-Dragonnet (Manhattan)}     & 198.31               &      -      &     -           \\
			{Dragonnet}                    & 200.89               & 0.994708    &  Fail to reject \\
			{NN-Dragonnet (Chebyshev)}     & 200.89               & 0.994708    &  Fail to reject \\
			{NN-Dragonnet (Euclidean)}     & 201.91               & 0.994708    &  Fail to reject \\
			\midrule
			{NN-TARnet (Euclidean)}        & 186.43               &      -      &     -           \\
			{NN-TARnet (Manhattan)}        & 190.16               & 0.841249    &  Fail to reject \\
			{NN-TARnet (Chebyshev)}        & 192.58               & 0.841249    &  Fail to reject \\
			{TARnet}                       & 232.83               & 0.013568    &  Reject \\
			\midrule
			{NN-NEDnet (Euclidean)}        & 175.99               &      -      &     -           \\
			{NN-NEDnet (Chebyshev)}        & 182.04               & 0.711365    &  Fail to reject \\
			{NN-NEDnet (Manhattan)}        & 191.46               & 0.468763    &  Fail to reject \\
			{NEDnet}                       & 252.51               & 0.000009    &  Reject \\
			\bottomrule
		\end{tabular}
	}
	
\end{table}

{Tables~\ref{Table:Synthetic2} presents the statistical analysis for the proposed causal inference models and their corresponding baseline models for Synthetic dataset, in terms of $\epsilon_{PEHE}$.
	Both statistical tests provide statistical evidence that the proposed NN-Dragonnet and NN-NEDnet 
	outperformed Dragonnet and NEDnet, respectively and are able to exhibit more reliable predictions.
	In contrast, although NN-TARnet present higher ranking than TARnet, there are no statistically significant differences in their performance, relative to $\epsilon_{PEHE}$ metric.}

\begin{table}
	\centering
	\setlength{\tabcolsep}{3pt}
	\renewcommand{\arraystretch}{1}
	\caption{FAR test and Finner post-hoc test based on  $\epsilon_{PEHE}$ for IHDP dataset\hfill\label{Table:Synthetic2}}
	{
		\begin{tabular}{lccc}
			\toprule
			\multirow{2}{*}{Model} &  \multirow{2}{*}{FAR} & \multicolumn{2}{c}{Finner post-hoc test}\\
			\cmidrule{3-4}                 &                       & $p_F$-value & $H_0$  \\
			\midrule
			{NN-Dragonnet (Manhattan)}    & 173.37               &      -      &     -           \\
			{NN-Dragonnet (Euclidean) }   & 177.45               & 0.802946    &  Fail to reject \\
			{NN-Dragonnet (Chebyshev)}    & 193.46               & 0.310031    &  Fail to reject \\
			{Dragonnet}                   & 257.72               & 0.000000    &   Reject \\
			\midrule
			{NN-TARnet (Manhattan)}       & 190.71               &      -      &     -           \\
			{NN-TARnet (Euclidean)}       & 199.64               & 0.630507    &  Fail to reject \\
			{TARnet}                      & 202.13               & 0.630507    &  Fail to reject \\
			{NN-TARnet (Chebyshev)}       & 209.51               & 0.578490    &  Fail to reject \\                  
			\midrule
			{NN-NEDnet (Chebyshev)}       & 155.34               &      -      &     -           \\
			{NN-NEDnet (Euclidean) }      & 172.58               & 0.291693    &  Fail to reject \\
			{NN-NEDnet (Manhattan)}       & 190.71               & 0.0045430   &  Reject \\
			{NEDnet}                      & 283.37               & 0.000000    &  Reject \\                     
			\bottomrule
		\end{tabular}
	}
	
\end{table}

\subsection{{Evaluation} on ACIC dataset}

\begin{figure*}[t]
	\subfigure[$\ \vert \epsilon_{ATE} \vert$]{\includegraphics[width=0.48\linewidth]{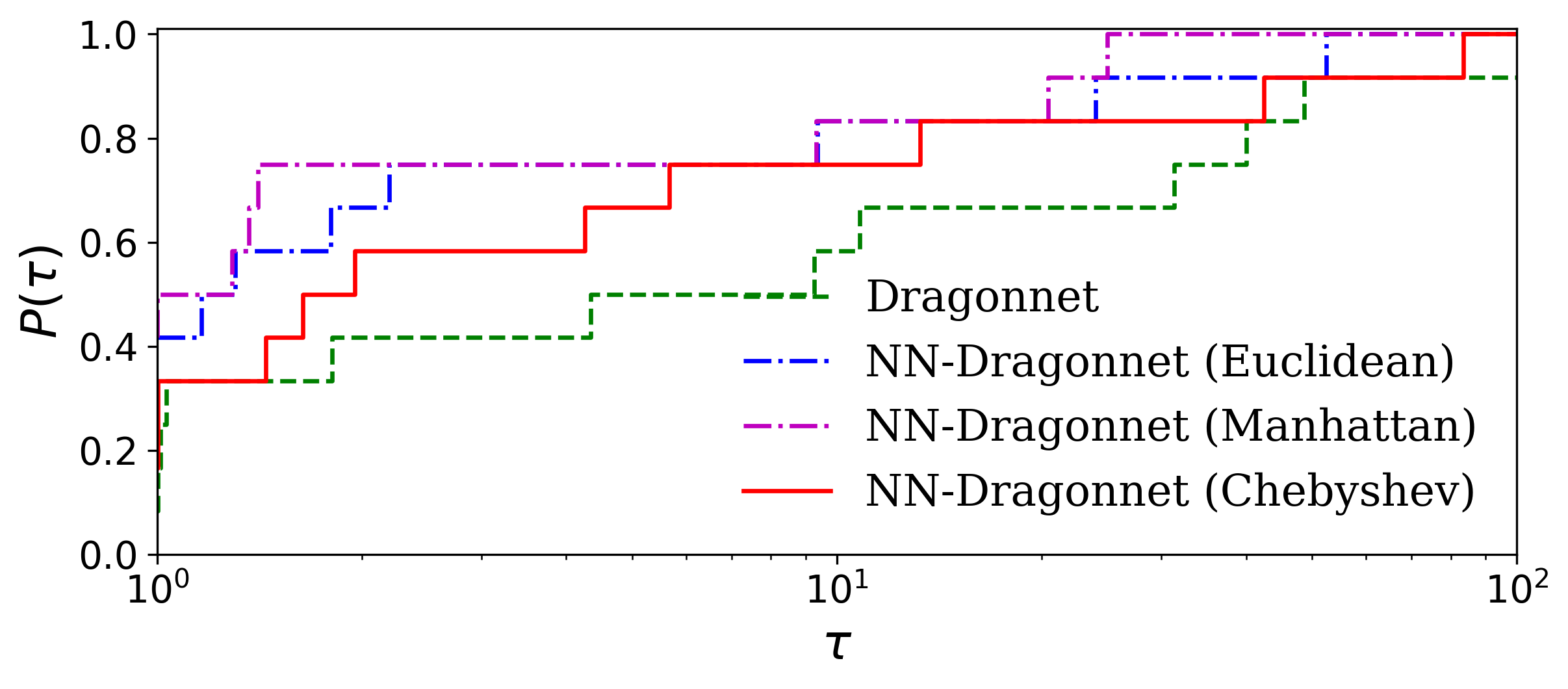}}\hfill
	\subfigure[$\ \epsilon_{PEHE} $]{\includegraphics[width=0.48\linewidth]{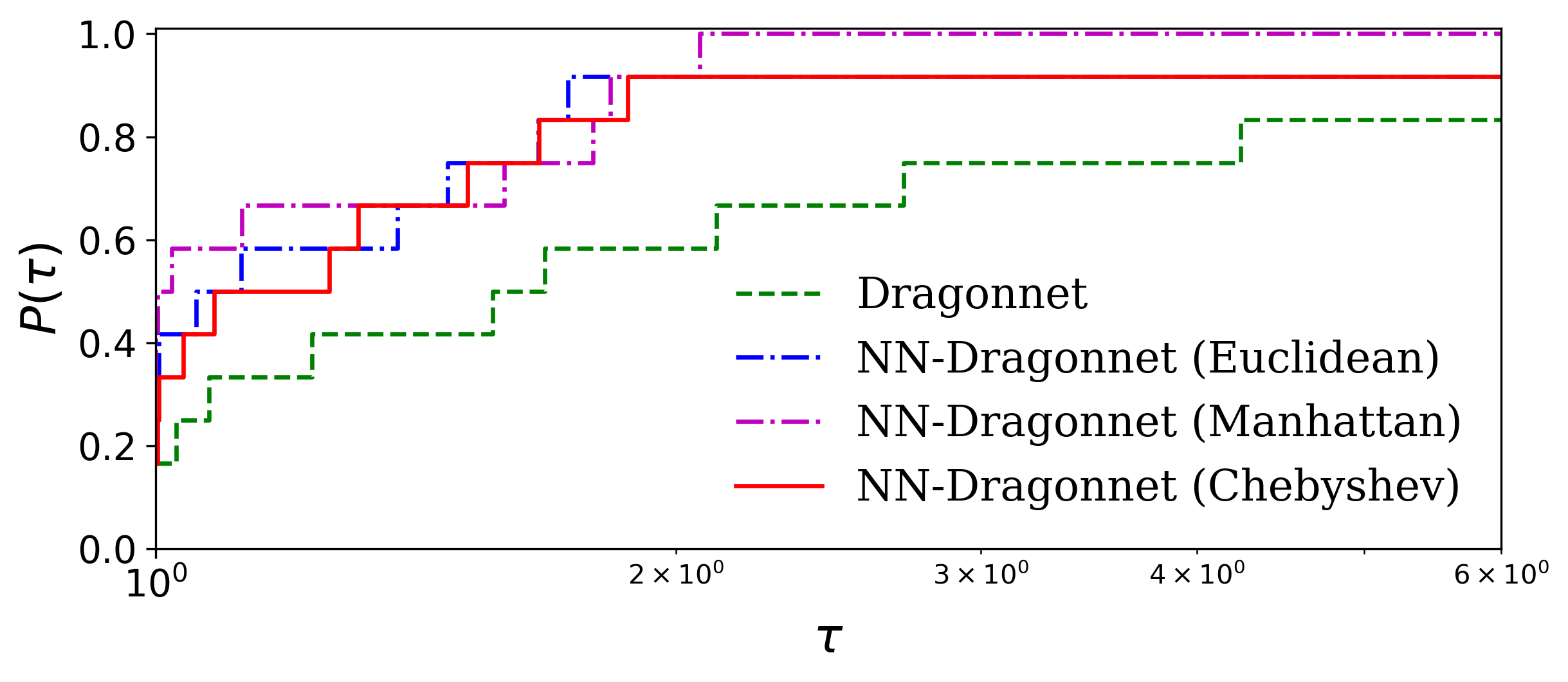}}
	\subfigure[$\ \vert \epsilon_{ATE} \vert$]{\includegraphics[width=0.48\linewidth]{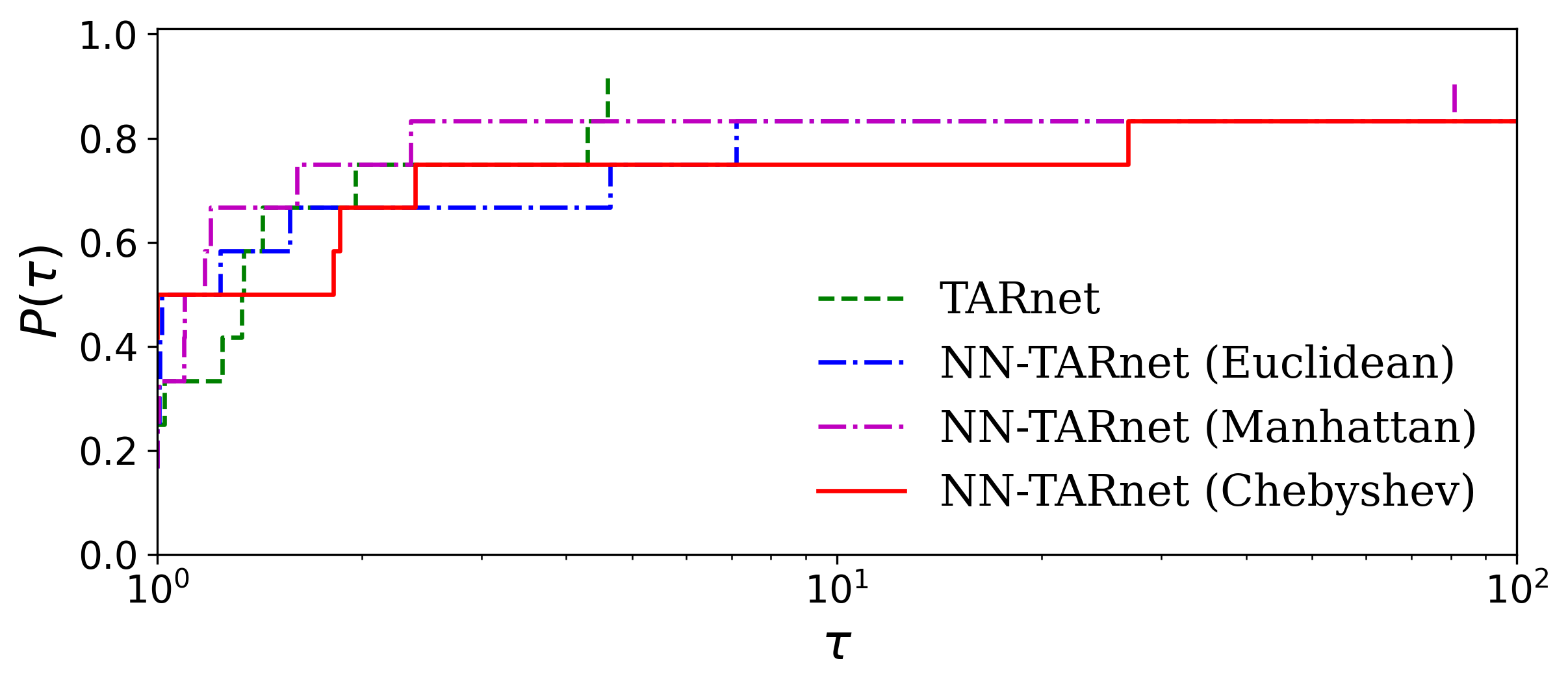}}\hfill
	\subfigure[$\ \epsilon_{PEHE} $]{\includegraphics[width=0.48\linewidth]{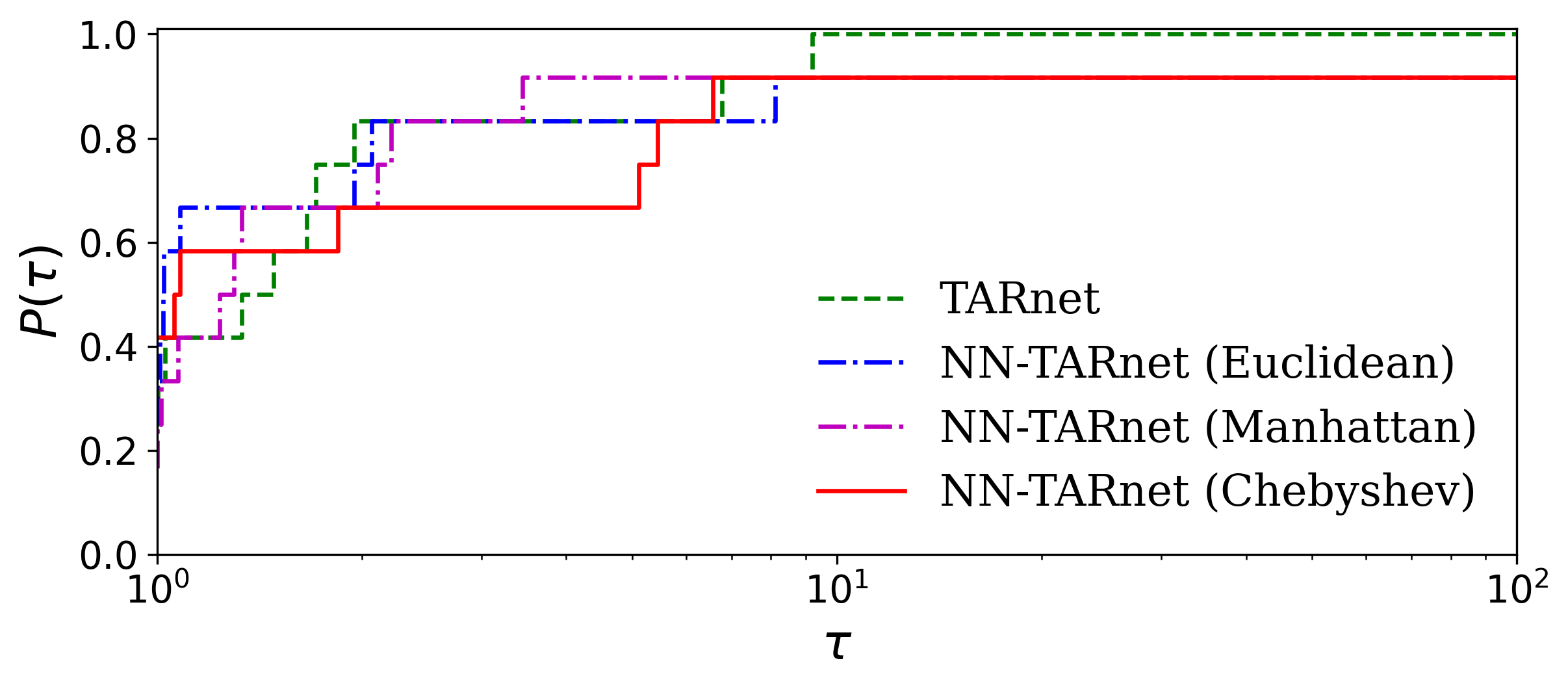}}
	\subfigure[$\ \vert \epsilon_{ATE} \vert$]{\includegraphics[width=0.48\linewidth]{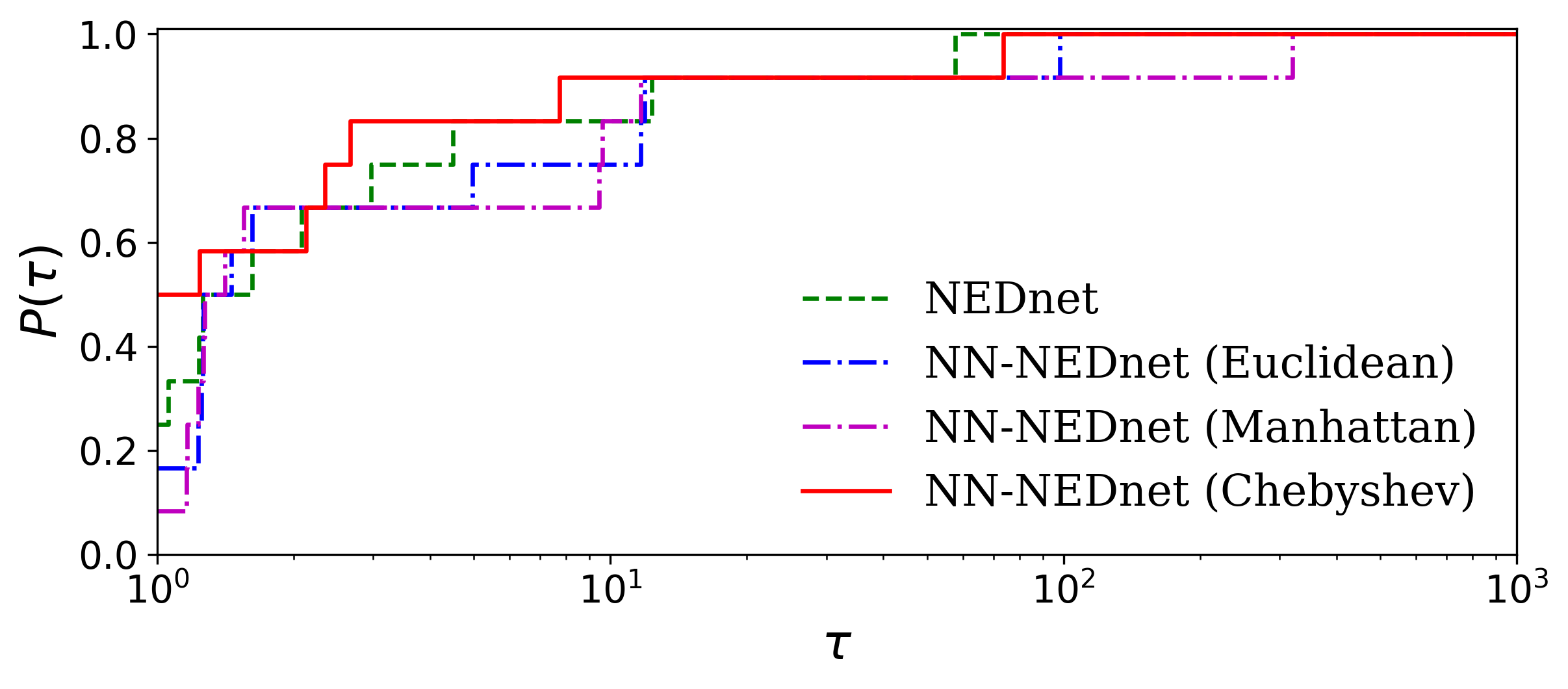}}\hfill
	\subfigure[$\ \epsilon_{PEHE} $]{\includegraphics[width=0.48\linewidth]{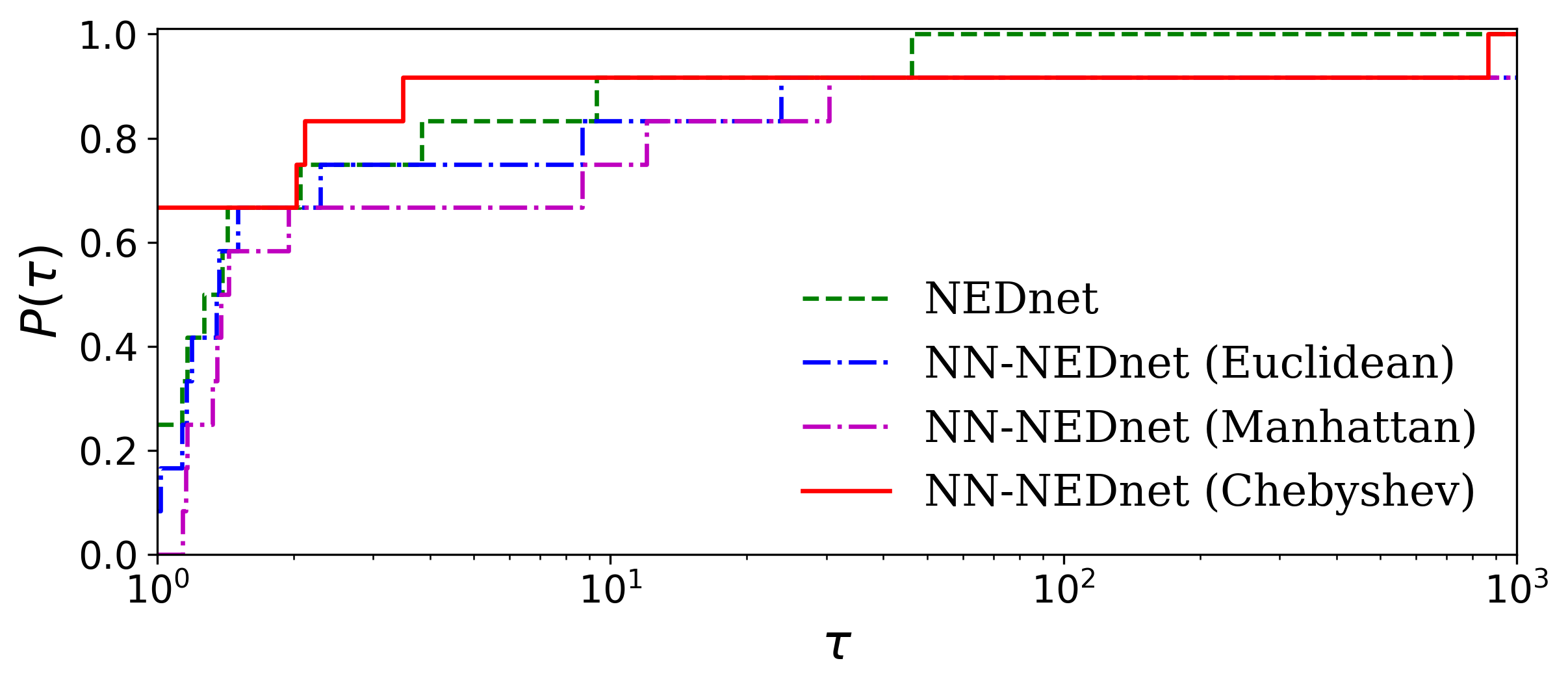}}
	\caption{Log$_{10}$ Performance profiles of models, based on $\vert \epsilon_{ATE} \vert $ and $\epsilon_{PEHE}$ for ACIC dataset} \label{fig:ACIC}
\end{figure*}

{Figure \ref{fig:ACIC}
	presents the performance profiles of NN-Dragonnet, NN-TARnet, NN-NEDnet and their corresponding
	baseline models, for $ \vert \epsilon_{ATE} \vert$ and $\epsilon_{PEHE}$ performance metrics.
	NN-Dragonnet outperform the state-of-the-art Dragonnet model independent of the used distance metric as regards both performance metrics. Additionally, it reported slightly better performance in case the Manhattan distance metric is used for the calculation of the neighboring instances.
	NN-TARnet (Chebyshev) model exhibits the best overall performance, slightly outperforming the rest of the models, relative to $\vert \epsilon_{ATE} \vert$, while NN-TARnet (Manhattan) reports the best performance, relative to $\epsilon_{PEHE}$.
	NN-NEDnet (Chebyshev) exhibits the highest probability of being the optimal model 
	in terms of effectiveness and robustness, since its curves lies on the top for both performance metrics.
	More specifically, NN-NEDnet solved 50\% and 66\% of the simulations with the best (lowest) $ \vert \epsilon_{ATE} \vert$ and $\epsilon_{PEHE}$ scores, while NEDnet reported only 23\% and 24\% in the same cases.}

{Tables~\ref{Table:ACIC1} and \ref{Table:ACIC2} present the statistical analysis for the proposed causal inference models and their corresponding baseline models for ACIC dataset, in terms of $ \vert \epsilon_{ATE} \vert$ and $\epsilon_{PEHE}$, respectively.
	NN-Dragonnet (Manhattan) is the top ranking model relative to both performance metrics. 
	FAR and Finner post-hoc tests suggests that NN-Dragonnet outperforms the baseline Dragonnet and there exhist statistically significant differences in their performances. Additionally, 
	NN-Dragonnet presents the best performance in case the Manhattan distance is utilized.
	Non-parametric FAR statistical test presents that NN-TARnet presents higher ranking compared to TARnet relative to both performance metrics; nevertheless, the Finner post-hoc test suggests that for all versions of NN-TARnet model there are no statistically significant differences in their performance.
	Finally, the interpretation of Tables~\ref{Table:ACIC1} and \ref{Table:ACIC2} suggest
	that NN-NEDnet presented slightly higher ranking than the baseline model NEDnet.}

\begin{table}
	\centering
	\setlength{\tabcolsep}{3pt}
	\renewcommand{\arraystretch}{1}
	\caption{FAR test and Finner post-hoc test based on $\vert \epsilon_{ATE} \vert $ for ACIC dataset\hfill\label{Table:ACIC1}}
	{
		\begin{tabular}{lccc}
			\toprule
			\multirow{2}{*}{Model} &  \multirow{2}{*}{FAR} & \multicolumn{2}{c}{Finner post-hoc test}\\
			\cmidrule{3-4}                 &                       & $p_F$-value & $H_0$  \\
			\midrule
			{NN-Dragonnet (Manhattan)}      & 18.25               &      -      &     -           \\
			{NN-Dragonnet (Euclidean)}      & 22.33               & 0.474959    &  Fail to reject \\
			{NN-Dragonnet (Chebyshev)}      & 26.58               & 0.209181    &  Fail to reject \\
			{Dragonnet}                     & 30.83               & 0.080796    &  Reject \\
			\midrule
			{NN-TARnet (Chebyshev)}         & 20.91               &      -      &     -           \\
			{NN-TARnet (Manhattan)}         & 22.75               & 0.812511    &  Fail to reject \\
			{NN-TARnet (Euclidean)}         & 23.33               & 0.812511    &  Fail to reject \\
			{TARnet}                        & 31.00               & 0.215446    &  Fail to reject \\
			\midrule
			{NN-NEDnet (Chebyshev)}         & 22.16               &      -      &     -           \\
			{NN-NEDnet (Euclidean)}         & 23.25               & 0.849667    &  Fail to reject \\
			{NN-NEDnet (Manhattan)}         & 25.41               & 0.763596    &  Fail to reject \\
			{NEDnet}                        & 27.16               & 0.763596    &  Fail to reject \\
			\bottomrule
		\end{tabular}
	}
	
\end{table}

\begin{table}
	\centering
	\setlength{\tabcolsep}{3pt}
	\renewcommand{\arraystretch}{1}
	\caption{FAR test and Finner post-hoc test based on  $\epsilon_{PEHE}$ for ACIC dataset\hfill\label{Table:ACIC2}}
	{
		\begin{tabular}{lccc}
			\toprule
			\multirow{2}{*}{Model} &  \multirow{2}{*}{FAR} & \multicolumn{2}{c}{Finner post-hoc test}\\
			\cmidrule{3-4}                 &                       & $p_F$-value & $H_0$  \\
			\midrule
			{NN-Dragonnet (Manhattan)}      & 19.25              &      -      &     -           \\
			{NN-Dragonnet (Euclidean) }     & 21.75              & 0.661815    &  Fail to reject \\
			{NN-Dragonnet (Chebyshev)}      & 23.58              & 0.590268    &  Fail to reject \\
			{Dragonnet}                     & 33.41              & 0.039045    &    Reject \\
			\midrule
			{NN-TARnet (Manhattan)}         & 20.75              &      -      &     -           \\
			{NN-TARnet (Euclidean)}         & 22.41              & 0.770588    &  Fail to reject \\
			{NN-TARnet (Chebyshev)}         & 24.58              & 0.649006    &  Fail to reject \\
			{TARnet}                        & 30.25              & 0.262418    &  Fail to reject \\                 
			\midrule
			{NN-NEDnet (Chebyshev)}         & 16.66              &      -      &     -           \\
			{NN-NEDnet (Euclidean)}         & 23.58              & 0.226216    &  Fail to reject \\
			{NEDnet}                        & 28.16              & 0.069785    &  Fail to reject \\
			{NN-NEDnet (Manhattan)}         & 29.58              & 0.069785    &    Fail to reject \\      
			\bottomrule
		\end{tabular}
	}
	
\end{table}

Summarizing, we point out that the proposed methodology significantly
improved the performance of the baseline models in case of imbalanced datasets
(IHDP), which indicates that the proposed approach is more beneficial to
challenging and complex benchmarks. This issue should be further taken into
consideration and analysis in order to study why and how the proposed NNIC
methodology improves the prediction accuracy especially in case of imbalanced problems.

{\section{Discussion}\label{sec:discussion}
	
	In this section, we provide a comprehensive discussion about 
	the motivation and contribution of this work as well as 
	about the advantages of the proposed NNCI methodology and its limitations.

	In a recent work, Kiriakidou and Diou \cite{KiriakidouAIAI22} proposed a new approach of 
	modifying the state-of-the-art causal inference model Dragonnet by enriching the model's inputs 
	with the average outcomes of the $k$-nearest neighbor samples from control and treatment groups. 
	Furthermore, the authors presented some promising numerical experiments using the IHDP dataset,
	which revealed the efficiency of their approach.
	However, the major drawbacks of this work were that the authors adopted 
	the proposed approach only on Dragonnet and evaluated the performance 
	of the proposed model only on one causal inference benchmark.

	Motivated by the efficiency of their approach and the promising experimental results, we extend the work conducted in \cite{KiriakidouAIAI22} and address their major drawbacks. 
	In this research, we propose NNCI methodology for integrating valuable information from the nearest neighboring samples from control and treatment groups in the training data, which is then utilized as inputs in the neural network models to provide accurate predictions of average and individual treatment effects. 
	The proposed NNCI methodology is then integrated to the most effective neural network-based causal inference models i.e. Dragonnet, TARnet and NEDnet. These models use a deep net to produce a representation layer, which is then processed independently for predicting both the outcomes for treatment and control groups. 
	
	The presented numerical experiments reveal that the application of NNCI
	considerably improve the performance of the state-of-the-art neural network models, achieving better estimation of both average and individual treatment effects. 
	The evaluation was performed using three of the most well-known causal inference benchmarks i.e. IHDP, Synthetic and ACIC, which are characterized by high complexity and challenge.
	The performance profiles as well as the Friedman Aligned-Ranks and Finner post-hoc tests provide strong statistical evidence about the effectiveness of the proposed
	approach. Therefore, based on this experimental analysis, we conclude that the
	proposed NNCI methodology in conjunction with modified versions of Dragonnet,
	TARnet and NEDnet is able to accurately estimate treatment effects.
	In more detail, as regards IHDP dataset, the proposed causal inference models outperformed the corresponding traditional ones for both $\vert \epsilon_{ATE} \vert $ and $\epsilon_{PEHE}$,
	independent of the utilized distance metric. It is worth mentioning that IHDP is an imbalanced dataset, which indicates that NNCI methodology proved beneficial for challenging and complex benchmarks.
	Concerning the Synthetic dataset, the experimental analysis reveals that the proposed models presented the best overall performance in case the Manhattan distance was used.
	Finally, regarding ACIC dataset, the adoption of the proposed NNCI methodology to the state-of-the-art causal inference models improved the average treatment effect estimation.

	It is worth mentioning that NNCI methodology can be adopted only to the selected neural network based causal inference models due to their special architecture design. More specifically, NNCI assumes the existence of a representation layer prior to the final treatment effect estimation layers.
	This can be considered as a limitation of the proposed work. Therefore, the adoption of the proposed methodology to other causal inference models
	models is an interesting research direction for future work.}

\section{Conclusion}\label{sec:conclusions}

In this work, we proposed a new methodology, named NNCI, which is applied to
well-established neural network-based models for the estimation of { individual (ITE) and average treatment effects (ATE)}. An advantage of the proposed methodology is the
exploitation of valuable information, not only from the covariates, but also
from the outcomes of nearest neighboring instances contained in the training
data from both treatment and control group.  {The proposed NNCI methodology is applied to the} state-of-the-art neural network models, Dragonnet, NEDnet and TARnet, { aiming to increase the models' prediction accuracy and reduce bias.}

The experimental analysis on three widely used datasets in the field of causal
inference, illustrated that the proposed approach improved the performance of
the traditional neural network-based models, regarding the estimation of causal
effects. This is confirmed by the performance profiles of Dolan and Mor{\'e}
as well as the nonparametric FAR test and the post-hoc Finner test. It is worth
highlighting that in all of the cases the proposed methodology leads to
considerable improvement in the estimation of treatment effects, in terms of
effectiveness and robustness.

Nevertheless, a limitation of the proposed work is the selection of the distance metric used
for calculating the nearest neighbors as well as the optimal value of parameter
$k$. An evaluation study on the effectiveness and sensitivity of different
values of parameter $k$ and distance metrics is included as future work.
{
	Another promising research subject can be
	considered the use of dynamic ensemble learning algorithms 
	\cite{alam2020dynamic,pintelas2020special,nandi2022reward,ortiz2016ensembles}, finite element machine learning \cite{amezquita2020machine,pereira2020fema} and self-supervised learning \cite{Adeli2023,hua2022uncertainty} for further improving the prediction 
	performance. Finally, an interesting idea for achieving more accurate predictions of causal effects is the development of new models for causal inference based on the architecture of augmented machine learning models \cite{bhattacharya2022epileptic,wu2022tuning,zhang2022enzymatic,zhang2022layered}.}

\section*{Aknowledgements} 

The work leading to these results has received funding from the European Union's Horizon 2020 research and innovation programme under Grant Agreement No. 965231, project REBECCA (REsearch on BrEast Cancer induced chronic conditions supported by Causal Analysis of multi-source data)


\end{document}